\documentclass[letterpaper]{article} 
\usepackage[preprint]{aaai2027}  
\usepackage[hyphens]{url}  
\usepackage{graphicx} 
\usepackage{natbib}  
\usepackage{caption} 
\usepackage{algorithm}
\usepackage{algorithmic}

\newcommand{\best}[1]{\textbf{#1}}

\usepackage{newfloat}
\usepackage{listings}
\DeclareCaptionStyle{ruled}{labelfont=normalfont,labelsep=colon,strut=off} 
\floatstyle{ruled}
\newfloat{listing}{tb}{lst}{}
\floatname{listing}{Listing}

\usepackage{booktabs}

\usepackage{multirow}
\usepackage{subcaption}
\usepackage{amsmath}
\usepackage{amsfonts}
\newtheorem{theorem}{Theorem}

\newtheorem{proposition}[theorem]{Proposition}

\title{AAAI Press Anonymous Submission\\Instructions for Authors Using \LaTeX{}}
\author{
    Written by AAAI Press Staff\textsuperscript{\rm 1}\thanks{With help from the AAAI Publications Committee.}\\
    AAAI Style Contributions by Peter Patel Schneider,
    Sunil Issar,\\
    J. Scott Penberthy,
    George Ferguson,
    Hans Guesgen,
    Francisco Cruz\equalcontrib\corresponding,
    Marc Pujol-Gonzalez\equalcontrib\corresponding
}
\affiliations{
    \textsuperscript{\rm 1}Association for the Advancement of Artificial Intelligence\\

    1101 Pennsylvania Ave, NW Suite 300\\
    Washington, DC 20004 USA\\
    proceedings-questions@aaai.org
}

\title{ActSafeGuard: Differentiable and Training-Aligned Constraint Enforcement for Flow-Matching Policies}
\author {
    Jianming Ma\equalcontrib\textsuperscript{\rm 1,\rm 2},
    Rongjun Jin\equalcontrib\textsuperscript{\rm 1},
    Xiaxi Si\textsuperscript{\rm 1},
    Yang Zhang\textsuperscript{\rm 1},
    Yiheng Li\textsuperscript{\rm 1},
    Yue Gao\textsuperscript{\rm 1, \rm 2}\corresponding
}
\affiliations {
    \textsuperscript{\rm 1}Shanghai Jiao Tong University\\
    \textsuperscript{\rm 2}Shanghai Innovation Institute\\
}

\begin{document}

\maketitle


\begin{abstract}

Vision-Language-Action (VLA) and World-Action Models (WAMs) have demonstrated strong capabilities in general-purpose robotic manipulation, yet their generated actions may violate hard physical constraints and therefore be unsafe or infeasible for deployment. Existing safety approaches either optimize statistical safety objectives without deterministic per-step guarantees or correct unsafe actions only during inference, creating a mismatch between policy training and execution. We introduce ActSafeGuard, a differentiable and training-aligned safeguard layer for flow-matching based policies. ActSafeGuard integrates hard action feasibility into policy learning, not merely treating safety as an inference-time external component. Through an analytical ray-scaling operator design, ActSafeGuard enables boundary-aware gradients to guide the model to naturally learn constrained manifolds. Extensive experiments on multiple standard foundation backbones ($\pi_{0.5}$ and Fast-WAM) across various tasks demonstrate that ActSafeGuard consistently achieves a $100\%$ step safety rate while fully preserving or even boosting task success rates, providing a scalable and minimally invasive solution for safe embodied AI deployment.

\end{abstract}


\section{Introduction}

Vision-Language-Action (VLA) models \cite{kim2024openvla,black2025pi05,nvidia2025gr00tn1openfoundation} and World-Action Models (WAMs) \cite{ye2026worldactionmodels,liao2025genieenvisioner} have recently shown remarkable progress in general-purpose robotic manipulation. By combining large-scale visual-language representations with generative action heads, these embodied foundation models can adapt to diverse downstream tasks while retaining broad semantic and motor priors. Despite their strong task-solving capabilities, however, the actions generated by these models are not always directly executable. Predicted actions may violate hard physical constraints, such as position limits, velocity bounds, or workspace restrictions, leading to unstable execution, hardware damage, or unsafe robot behavior.

For robotic manipulation, safety must be enforced at every execution step, because a single infeasible action can result in an unacceptable physical consequence. This is particularly problematic for flow-matching action heads, which generate action chunks in an unconstrained space without explicit feasibility guarantees. Thus, a practical safeguard must provide deterministic per-step feasibility while preserving the task competence and generative flexibility of pretrained foundation policies.

\begin{figure}
    \centering
    \includegraphics[width=\linewidth]{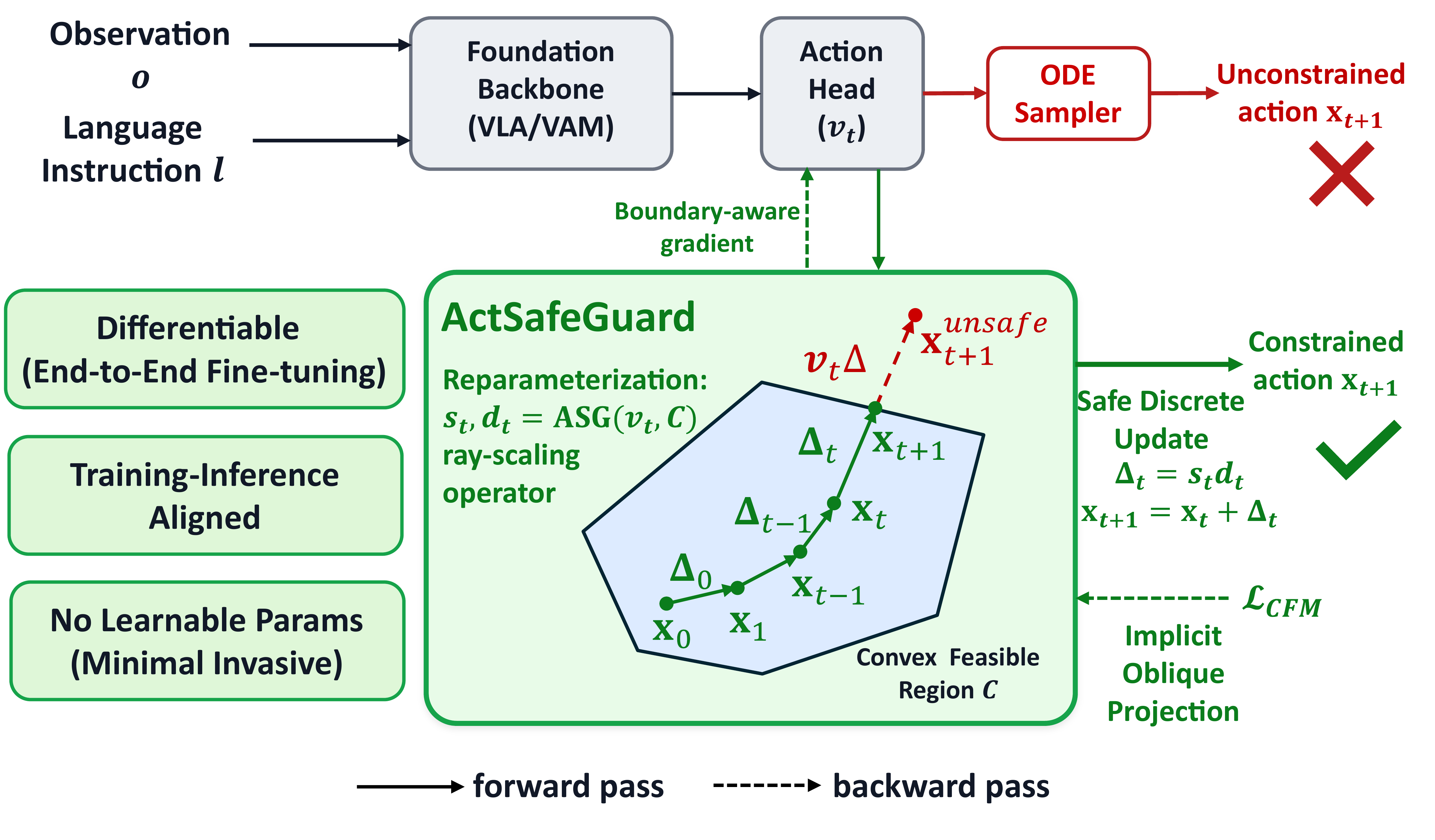}
    \caption{\textbf{Overview of ActSafeGuard.} Given an observation and a language instruction, a foundation VLA/WAM backbone and its flow-matching action head predict an unconstrained velocity $v_t$. ActSafeGuard transforms $v_t$ into an update direction $d_t$ and a safe scaling factor $s_t$, adaptively rescaling constraint-violating steps such that the trajectory remains within the convex feasible region. During backpropagation, its gradient transformation admits an implicit oblique-projection interpretation, providing boundary-aware learning signals.}
    \label{fig:overall}
\end{figure}

Existing safety approaches generally address this challenge from either the training side or the inference side. Training-time constrained methods incorporate safety costs \cite{zhang2026safevla} or regularization objectives \cite{wei2026can} into policy optimization, allowing the policy to become safety-aware but typically providing only probabilistic or expectation-based guarantees. In contrast, inference-time safeguards directly correct infeasible actions through projection, clipping, or optimization-based filtering \cite{hu2025vlsa}. Although these methods can enforce hard constraints during deployment, the correction mechanism is absent from policy training. Consequently, the policy is optimized in an unconstrained action space but executed under an externally modified action distribution, introducing a training--inference mismatch that may distort generated trajectories and degrade task performance. The central limitation is therefore not simply the absence of a safety mechanism, but the separation of hard constraint enforcement from policy learning.

In this work, we introduce ActSafeGuard, a differentiable and training-aligned constraint operator for flow-matching-based policies. Our key insight is that \emph{hard action feasibility can be integrated directly into the policy learning process, rather than being treated merely as an external inference-time correction.} ActSafeGuard is inserted at the output of the flow-matching action head and applies an analytical ray-scaling operator to each discrete flow update. 
Starting from a feasible initial point, this construction ensures that every constrained flow step remains within the state-dependent convex feasible set.
Crucially, the same safeguard operator is active during both training and inference. 
During training, its differentiable construction allows gradients to propagate through the constrained update, enabling the action head to receive boundary-aware learning signals. This training-aligned design encourages the policy to adapt its predicted directions to the geometry of the feasible action space, thereby preserving effective learning while enforcing hard feasibility. Moreover, ActSafeGuard introduces no additional learnable network and can be incorporated into pretrained flow-matching action heads with minimal modification.

We evaluate ActSafeGuard on representative foundation backbones across multiple manipulation tasks with various constraints. The results show that ActSafeGuard consistently achieves a $100\%$ step safety rate while preserving task success rates compared with other constrained methods. Our main contributions are summarized as follows:

\begin{itemize}
\item We propose ActSafeGuard, a differentiable and training-aligned constraint operator that integrates hard action feasibility directly into the learning and generation processes of flow-matching based policies.

\item We show that ActSafeGuard induces boundary-aware gradients through an implicit oblique projection, enabling constrained updates to slide along feasible boundaries instead of being merely truncated.

\item Extensive experiments with $\pi_{0.5}$ and Fast-WAM demonstrate deterministic constraint satisfaction while preserving task success and action-generation quality.

\end{itemize}

\section{Related Works}

\paragraph{Safety Constrained VLA and WAMs.}
Safety is a core desideratum for VLA and WAM policies deployed in open-ended environments, where unconstrained physical actions may lead to catastrophic failures \cite{li2026embodiedaisafety,li2026vlasafety,kim2026safeembodied,kim2026modularguardrails}. Existing methods mainly address this issue from either the training side or the inference side:

\textit{(1) Training-time Constrained Policies.} 
One paradigm embeds safety directly into the policy optimization process \cite{zhang2026safevla, tang2026safedojo, wei2026can}. For instance, SafeVLA \cite{zhang2026safevla} formulates the safety task as a constrained MDP and leverages safe reinforcement learning (RL) to achieve VLA safety alignment. SafeDojo \cite{tang2026safedojo} utilizes a generative video world model to produce imaginary rollouts and introduces a Lagrangian-constrained GRPO objective for policy training. These methods can improve safety awareness, but they typically optimize expected costs or soft penalties and therefore do not guarantee deterministic per-step feasibility. Furthermore, incorporating constrained optimization techniques (e.g., Lagrangian multipliers) into the training pipeline introduces a non-trivial trade-off between task objective and constraint satisfaction.

\textit{(2) Inference-time Safety Guards.} 
Inference-time safeguards instead modify generated actions during deployment. One branch of work \cite{hu2025vlsa, english2026neurosymbolic} constructs control barrier functions (CBFs) and safety sets to project actions by solving a Quadratic Program (QP) optimization at each sampling step, ensuring that the generated actions strictly reside within the feasible region.
Another branch adopts action masking \cite{beaudin2026anybodyguard} or heuristic-based hard truncation \cite{chandra2026physvla} to filter out unsafe actions. 
While such methods can enforce hard constraints at execution time, they are absent from policy training, creating a training--inference mismatch that may distort the learned action distribution and degrade task performance.

In summary, training-time constrained methods simultaneously optimize for task competence and safety regularizers but fall short of delivering deterministic constraint satisfaction. Conversely, inference-time safety guards mathematically enforce hard constraints during deployment but introduce a training-inference mismatch that often compromises the backbone's execution capability. How to effectively balance task performance with strict constraint satisfaction through a training-inference aligned framework remains an open challenge.

\section{Preliminary}

\paragraph{Policy Formulation.}
We formulate language-conditioned robotic manipulation as learning a mapping from multimodal observations to a sequence of future actions. At control step $t$, the observation tuple is defined as $\mathbf{o}_t = [\{\mathbf{I}_t^i\}_{i=1}^n, \mathbf{q}_t]$, comprising multi-view images and proprioceptive states. Guided by a language instruction $l$, a VLM/VGM backbone $f_\theta$ first projects the inputs into a unified latent representation:
$\mathbf{e}_t = f_\theta(\mathbf{o}_t, l).$
An action head $\pi_\phi$ then decodes this embedding into an action chunk $\mathbf{a}_{t:t+H-1}$, representing motor commands over a future horizon $H$. The policy is composed as $\Pi(\mathbf{o}_t, l) = \pi_\phi(\mathbf{e}_t)$, and the generated chunk is executed via a receding-horizon scheme.

\paragraph{Flow-Matching Action Generation.} 
Flow matching \cite{lipman2023flow} is a prevalent paradigm for training the generative action head $\pi_\phi$. Let $\mathbf{x}_1$ denote the target expert action chunk with horizon H (i.e., $\mathbf{a}_{t:t+H-1}$) and $\mathbf{x}_0 \sim p(\mathbf{x}_0)$ denote an initial noise sample. Flow matching constructs a probability path via linear interpolation:
\begin{equation}
\mathbf{x}_\tau = (1-\tau)\mathbf{x}_0 + \tau \mathbf{x}_1, \quad \tau \in [0,1],
\end{equation}
where $\tau$ denotes the continuous flow timestep. The action head, instantiated as a velocity field network $v_\phi(\mathbf{x}_\tau, \tau, \mathbf{e}_t)$ conditioned on the latent context $\mathbf{e}_t$, is trained to approximate the target vector field $u^\star = \mathbf{x}_1 - \mathbf{x}_0$. The training objective is:
\begin{equation}
\mathcal{L}_{\mathrm{CFM}} = \mathbb{E}_{\mathbf{x}_0, \mathbf{x}_1, \tau} \left[ \left\| v_\phi(\mathbf{x}_\tau, \tau, \mathbf{e}_t) - (\mathbf{x}_1 - \mathbf{x}_0) \right\|_2^2 \right].
\label{eq:cfm}
\end{equation}
During inference, starting from $\mathbf{x}_0$, an Ordinary Differential Equation (ODE) solver integrates the learned vector field over $\tau \in [0, 1]$ to synthesize the final action chunk $\mathbf{x}_1$.

\paragraph{Constrained Action Space.} 
In real-world robotic applications, physical and environmental limitations impose strict constraints on the action space. We characterize the feasible region as a state-dependent convex polytope:
\begin{equation}
\mathcal{C}(\mathbf{o}_t) = \{\mathbf{x}^c \mid \mathbf{A}(\mathbf{o}_t)\mathbf{x}^c \leq \mathbf{b}(\mathbf{o}_t)\},
\end{equation}
where $\mathbf{x}^c$ denotes the dimensions of the action chunk subject to hard constraints, and $\mathbf{A}(\mathbf{o}_t), \mathbf{b}(\mathbf{o}_t)$ define observation-conditioned boundaries. Our goal is to ensure that $\mathbf{x}^c$ strictly resides within $\mathcal{C}(\mathbf{o}_t)$ throughout the sampling process, thereby guaranteeing deterministic, zero-violation execution upon deployment.

\section{Method}

To achieve zero-constraint-violation generation with minimal intervention to pretrained models, we propose \textbf{ActSafeGuard}. The core idea is to embed a differentiable, parameter-free ray-scaling operator directly into a discrete-time flow-matching process, ensuring per-step feasibility while facilitating boundary-aware gradient updates during training.

\subsection{Discrete-Time Flow Matching with Feasible States}
Standard flow matching integrates a continuous vector field over time. However, verifying continuous-time safety analytically is often intractable. Instead, ActSafeGuard formulates the generation as an $N$-step discrete flow process. Let $\tau_k = k/N$ for $k \in \{0, \dots, N-1\}$. The exact probability path is discretized as:
$
    \mathbf{x}_k = (1-\tau_k)\mathbf{x}_0 + \tau_k \mathbf{x}_1,
$
where the ideal one-step update is given by $\Delta^\star = \mathbf{x}_{k+1} - \mathbf{x}_k = (\mathbf{x}_1-\mathbf{x}_0)/N$. 
Instead of learning the continuous velocity $v_\phi$, the action head is parameterized to directly predict the discrete step $\Delta_\phi(\mathbf{x}_k, \tau_k, \mathbf{e}_t)$. The model can be trained using a discrete-time flow matching loss:
\begin{equation}
    \mathcal{L}_{\mathrm{DFM}} = \mathbb{E}_{k, \mathbf{x}_0, \mathbf{x}_1} \left[ \left\| \Delta_\phi(\mathbf{x}_k, \tau_k, \mathbf{e}_t) - \Delta^\star \right\|_2^2 \right].
    \label{eq:dfm}
\end{equation}
The formulation of Eq.~(\ref{eq:dfm}) is mathematically aligned with the CFM objective in Eq.~(\ref{eq:cfm}) used by standard continuous flow matching. Although discretization introduces bounded approximation error~\cite{ma2026polyflow}, it preserves the vector-field prior of continuous pretrained backbones by interpreting their predicted velocity as a nominal discrete update direction. The advantage of this formulation is that feasibility can be enforced step by step: if $\mathbf{x}_0 \in \mathcal{C}(\mathbf{o}_t)$ and every constrained update respects the boundary, then the entire generated sequence $\mathbf{x}_0 \rightarrow \dots \rightarrow \mathbf{x}_N$ remains feasible.

\subsection{ActSafeGuard Parameterization}

Given the discrete formulation above, ActSafeGuard enforces feasibility by modifying each nominal flow update before it is applied. The design goal is to preserve the vector-field prior of the pretrained action head as much as possible: the predicted velocity still determines the update direction, while ActSafeGuard only rescales its magnitude when the update would leave the feasible set.

For notational simplicity, we describe the operation on the constrained action dimensions. At step $k$, the pretrained action head predicts an unconstrained velocity
$\mathbf{v}_\phi(\mathbf{x}_k,\tau_k,\mathbf{e}_t)$. We convert it into the following two components:

\paragraph{Direction Vector:} This vector dictates the intended moving direction along with its step magnitude:
\begin{equation}
    \mathbf{d}_\phi = \frac{\mathbf{v}_\phi(\mathbf{x}_k, \tau_k, \mathbf{e}_t)}{N},
\end{equation}
where $N$ denotes the total number of discretization steps. 

\paragraph{Safety Scaling Factor:} To guarantee that the next state remains within the closed feasible domain, we adaptively constrain the magnitude of $\mathbf{d}_\phi$ using a safety weight $s_\phi \in [0, 1]$. ActSafeGuard modulates the nominal step to yield the final safe discrete step:
\begin{equation}
    \Delta_\phi = s_\phi \mathbf{d}_\phi.
\end{equation}

\paragraph{Computation of $s_\phi$ via Ray Shooting.} Given the direction vector $\mathbf{d}_\phi$ and the current state $\mathbf{x}_k$, the ray-shooting operator computes the intersection $\mathbf{z}$ with the closest constraint boundary along the line of travel:
\begin{equation}
    \alpha = \min_{i: \mathbf{a}_i^\top \mathbf{d}_\phi > 0} \frac{b_i - \mathbf{a}_i^\top \mathbf{x}_k}{\mathbf{a}_i^\top \mathbf{d}_\phi}, \quad \mathbf{z} = \mathbf{x}_k + \alpha \mathbf{d}_\phi,
    \label{eq:alpha}
\end{equation}
where $\mathbf{a}_i^\top$ and $b_i$ denote the $i$-th row and element of the constraint matrix $\mathbf{A}$ and boundary vector $\mathbf{b}$ respectively, and $\alpha \geq 0$ is a scalar representing the maximum allowable scaling factor before hitting the nearest constraint facet $i$. Consequently, the safety scaling factor $s_\phi$ is defined as:
\begin{equation}
    s_\phi = \min(1, \alpha) = 
    \begin{cases}
        1, & \alpha \geq 1, \\
        \alpha, & \alpha < 1.
    \end{cases}
    \label{eq:s}
\end{equation}
This piecewise formulation yields two regimes:
\begin{itemize}
    \item When $\alpha \geq 1$, the nominal update $\mathbf{x}_k + \mathbf{d}_\phi$ is already feasible and ActSafeGuard leaves it unchanged.
    \item When $\alpha < 1$, the nominal update would cross a boundary, so ActSafeGuard shortens it exactly to the ray-boundary intersection:
    \begin{equation}
        \mathbf{x}_{k+1}
        =
        \mathbf{x}_k+\Delta_\phi
        =
        \mathbf{x}_k+\alpha\mathbf{d}_\phi
        =
        \mathbf{z}\in\partial\mathcal{C}.
    \end{equation}
\end{itemize}


\begin{figure}
    \centering
    \includegraphics[width=0.9\linewidth]{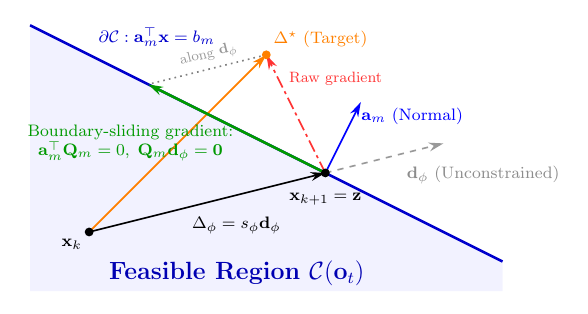}
    \caption{\textbf{ActSafeGuard update and boundary-sliding gradient.}
When the nominal update $\mathbf{d}_\phi$ crosses the active constraint facet
$\mathbf{a}_m^\top \mathbf{x}=b_m$, ActSafeGuard rescales the step to the boundary
intersection $\mathbf{z}$. Around this boundary-limited point, the local Jacobian
acts as an oblique projection $\mathbf{Q}_m$, satisfying
$\mathbf{a}_m^\top\mathbf{Q}_m=0$ and $\mathbf{Q}_m\mathbf{d}_\phi=\mathbf{0}$.
Thus, raw gradients that would otherwise point outside the feasible region are
converted into boundary-sliding gradients along the active facet, encouraging the
policy to re-orient unsafe updates while preserving hard feasibility. }
    \label{fig:placeholder}
\end{figure}

\begin{proposition}[Safety Guarantee]
    By construction, ActSafeGuard guarantees that the generated trajectory remains within the closed feasible region $\mathcal{C}(\mathbf{o}_t)$ across all inference steps, provided that the initial noise satisfies $\mathbf{x}_0 \in \mathcal{C}(\mathbf{o}_t)$. 
\end{proposition}

\begin{proposition}[Differentiability]
A key advantage of ActSafeGuard is its end-to-end differentiability, allowing it to be embedded directly into training loops without additional modification. The complete forward pass for the ActSafeGuard can be expressed as:
\begin{equation}
    \Delta_\phi = \min\Big(1, \min_{i: \mathbf{a}_i^\top \mathbf{v}_\phi > 0} \frac{b_i - \mathbf{a}_i^\top \mathbf{x}_k}{\mathbf{a}_i^\top \mathbf{v}_\phi} N\Big)\cdot \frac{\mathbf{v}_\phi}{N},
\end{equation}
where $\mathbf{v}_\phi$ is the shortcut notation for the velocity field prediction $\mathbf{v}_\phi(\mathbf{x}_k, \tau_k, \mathbf{e}_t)$. The gradient of the output $\Delta_\phi$ with respect to the input $\mathbf{v}_\phi$ exists in closed-form everywhere except at the boundary corners, which can be handled via standard subgradient methods during backpropagation.
\end{proposition}

The training and inference procedures of ActSafeGuard are summarized in Algorithm \ref{alg:training} and Appendix~G.

\begin{algorithm}[h]
\caption{ActSafeGuard Training}\label{alg:training}
\begin{algorithmic}[1]
\REQUIRE Pretrained policy parameters $\phi$, constraint sets $\mathcal{C}(\mathbf{o}_t)$
\FOR{each batch $(\mathbf{o}_t, \mathbf{x}_1)$ guided by instruction $l$}
    \STATE Encode context: $\mathbf{e}_t = f_\theta(\mathbf{o}_t, l)$
    \STATE Sample step $k \sim \mathcal{U}(0, N-1)$ and initial noise $\mathbf{x}_0 \sim p(\mathbf{x}_0)$ s.t. $\mathbf{x}_0 \in \mathcal{C}(\mathbf{o}_t)$
    \STATE Compute interpolated state $\mathbf{x}_k = (1 - \frac{k}{N})\mathbf{x}_0 + \frac{k}{N} \mathbf{x}_1$
    \STATE Predict baseline velocity $\mathbf{v}_\phi \leftarrow \mathbf{v}_\phi(\mathbf{x}_k, \tau_k, \mathbf{e}_t)$
    \STATE Compute direction vector $\mathbf{d}_\phi = \mathbf{v}_\phi / N$
    \STATE Compute scale $s_\phi$ using Eq (\ref{eq:alpha}) and (\ref{eq:s})
    \STATE Apply safe update $\Delta_\phi = s_\phi \mathbf{d}_\phi$
    \STATE Compute discrete flow matching loss $\mathcal{L}_{\text{DFM}} = \| \Delta_\phi - (\mathbf{x}_1 - \mathbf{x}_0)/N \|_2^2$
    \STATE Update policy parameters $\phi \leftarrow \phi - \eta \nabla_\phi \mathcal{L}_{\text{DFM}}$
\ENDFOR
\end{algorithmic}
\end{algorithm}

\subsection{Gradient-Based Boundary Sliding Correction}

We now examine why ActSafeGuard provides boundary-aware learning signals rather than behaving as a non-differentiable clipping module. The key lies in the local Jacobian of the safe update $\Delta_\phi$ with respect to the nominal direction $\mathbf{d}_\phi$.

Consider the boundary-limited case $\alpha<1$, and assume that the active facet is locally unique:
\begin{equation}
    \hat{i} =
    \arg\min_{i:\mathbf{a}_i^\top \mathbf{d}_\phi>0}
    \frac{b_i-\mathbf{a}_i^\top \mathbf{x}_k}
    {\mathbf{a}_i^\top \mathbf{d}_\phi}.
\end{equation}
Let $c_{\hat{i}}=b_{\hat{i}}-\mathbf{a}_{\hat{i}}^\top\mathbf{x}_k$. In this regime, the safe update is
\begin{equation}
    \Delta_\phi
    =
    \alpha\mathbf{d}_\phi
    =
    \frac{c_{\hat{i}}\mathbf{d}_\phi}
    {\mathbf{a}_{\hat{i}}^\top\mathbf{d}_\phi}.
\end{equation}
Taking the local differential with respect to $\mathbf{d}_\phi$ gives
\begin{equation}
    \mathrm{d}\Delta_\phi
    =
    \alpha \mathbf{Q}_{\hat{i}}\,\mathrm{d}\mathbf{d}_\phi,
    \qquad
    \mathbf{Q}_{\hat{i}}
    =
    \mathbf{I}
    -
    \frac{\mathbf{d}_\phi\mathbf{a}_{\hat{i}}^\top}
    {\mathbf{a}_{\hat{i}}^\top\mathbf{d}_\phi}.
\end{equation}
The matrix $\mathbf{Q}_{\hat{i}}$ is an oblique projection onto the tangent space of the active facet along the ray direction. It satisfies
\begin{equation}
    \mathbf{a}_{\hat{i}}^\top \mathbf{Q}_{\hat{i}}=\mathbf{0}^\top,
    \qquad
    \mathbf{Q}_{\hat{i}}\mathbf{d}_\phi=\mathbf{0}.
\end{equation}
Therefore, once the nominal update reaches a boundary, first-order changes in the safe update cannot move outward through the active constraint. Instead, the Jacobian preserves only variations that slide the boundary intersection along the feasible facet.

This property explains the boundary-sliding behavior of ActSafeGuard. Unlike hard clipping or stop-gradient correction, the ray-scaling operator remains differentiable in the boundary-limited regime. Gradients from the training objective can still flow through $\Delta_\phi$ to $\mathbf{d}_\phi$, but they are geometrically reshaped by $\mathbf{Q}_{\hat{i}}$. As a result, the action head receives learning signals that encourage it to re-orient unsafe nominal directions along the feasible boundary, rather than merely increasing a step magnitude that would be truncated by the safeguard.

\begin{table*}[t]
\centering
\begin{tabular}{ll ccccc c ccccc}
\toprule
\multirow{2}{*}{\textbf{Backbone}}
& \multirow{2}{*}{\textbf{Method}}
& \multicolumn{5}{c}{\textbf{Static Constraints (PosCons)}}
&
& \multicolumn{5}{c}{\textbf{Dynamic Constraints (PosCons + VelCons)}} \\
\cmidrule(lr){3-7} \cmidrule(lr){9-13}
&
& \textit{lp} & \textit{ps} & \textit{hm} & \textit{pec} & \textbf{Mean SR}
&
& \textit{lp} & \textit{ps} & \textit{hm} & \textit{pec} & \textbf{Mean SR} \\
\midrule

\multirow{5}{*}{$\pi_{0.5}$}
& Baseline
& 100 & 90 & 21 & 90 & 75.25
&
& 100 & 90 & 21 & 90 & 75.25 \\

& Projection
& \best{100} & 90 & 21 & 88 & 74.75
&
& \best{100} & 84 & 26 & 95 & 76.25 \\

& Truncation
& 99 & \best{91} & 27 & 92 & 77.25
&
& 99 & 90 & 27 & 91 & 76.75 \\

& GaugeFlow
& \best{100} & 82 & 25 & 93 & 75.00
&
& N/A & N/A & N/A & N/A & N/A \\

& \textbf{ActSafeGuard}
& \best{100} & 90 & \best{34} & \best{97} & \best{80.25}
&
& \best{100} & \best{98} & \best{31} & \best{97} & \best{81.50} \\

\midrule

\multirow{4}{*}{Fast-WAM}
& Baseline
& 100 & 93 & 42 & 98 & 83.25
&
& 100 & 93 & 42 & 98 & 83.25 \\

& Projection
& \best{100} & 85 & 40 & 80 & 76.25
&
& \best{100} & 85 & 35 & 82 & 75.50 \\

& Truncation
& 98 & 80 & \best{50} & 90 & 79.50
&
& \best{100} & 94 & \best{45} & 88 & 81.75 \\

& GaugeFlow
& 100 & 80 & 39 & 90 & 77.25
&
& N/A & N/A & N/A & N/A & N/A \\

& \textbf{ActSafeGuard}
& \best{100} & \best{90} & 45 & \best{93} & \best{82.00}
&
& 97 & \best{95} & \best{45} & \best{95} & \best{83.00} \\

\bottomrule
\end{tabular}
\caption{\textbf{Quantitative task success performance (SR, \%) across all evaluation scenarios.}
We evaluate methods across four manipulation tasks: \textit{lift pot} (\textit{lp}),
\textit{place shoe} (\textit{ps}), \textit{hanging mug} (\textit{hm}), and
\textit{place empty cup} (\textit{pec}), under static position constraints
(\textbf{PosCons}) and dynamic position + velocity constraints
(\textbf{PosCons} + \textbf{VelCons}). 
Step Safety Rate (SSR) is omitted as all constrained methods deterministically
achieve 100\% SSR. Bold numbers indicate the best performance among constrained methods. Training and evaluation details are in Appendix~D and E.}
\label{tab:main_results_combined_highlight}
\end{table*}

\section{Experiments}
\label{sec:experiments}

We conduct experiments to evaluate whether ActSafeGuard can enforce hard action feasibility without sacrificing the task competence of pretrained flow-matching policies. The experiments are organized around three research questions:
\begin{itemize}
    \item \textbf{RQ1: Performance preservation.} Can ActSafeGuard eliminate constraint violations while maintaining the task success rate of the original policy?
    \item \textbf{RQ2: Generalizability to complex constraints.} Can ActSafeGuard handle high-dimensional, state-dependent, and multi-step dynamic constraints?
    \item \textbf{RQ3: Differentiability.} How much does the differentiable ray-scaling design contribute to stable constrained policy learning?
\end{itemize}

\subsection{Experimental Setup}
\label{sec:exp_setup}

\paragraph{Simulation environment and tasks.}
We evaluate ActSafeGuard in RoboTwin~\cite{chen2025robotwin} with the bimanual ALOHA-Agilex platform. The robot has two 6-DoF arms and two grippers, yielding a 14-dimensional action space. We consider four manipulation tasks of different difficulty and contact structure: \emph{lift pot}, \emph{place shoe}, \emph{hanging mug}, and \emph{place empty cup}. 

\paragraph{Policy input and output.}
The policy uses absolute joint and gripper positions as actions. At each control step $t$, it receives a multimodal observation
$\mathbf{o}_t = [\{\mathbf{I}_t^i\}_{i=1}^{3}, \mathbf{q}_t, \mathbf{g}_t],$
where $\{\mathbf{I}_t^i\}_{i=1}^{3}$ are three camera views, $\mathbf{q}_t \in \mathbb{R}^{12}$ denotes the current arm-joint positions, and $\mathbf{g}_t \in \mathbb{R}^{2}$ denotes the gripper states. Given a language instruction $l$, the policy predicts an action chunk $\mathbf{a}_{t:t+H-1} = [\mathbf{a}_t,\dots,\mathbf{a}_{t+H-1}]$ with horizon $H$.

\paragraph{Safety constraints.}
We impose hard constraints on the 12-DoF arm-joint subspace, denoted by the superscript $q$, while leaving the gripper dimensions unconstrained. Detailed parameterizations are provided in Appendix~A.

\textbf{Position constraints (PosCons)} define a static joint-limit box:
\begin{equation}
    \mathbf{q}_{\min} \leq \mathbf{a}_{t+i}^{q} \leq \mathbf{q}_{\max}, 
    \quad \forall i \in \{0,\dots,H-1\}.
\end{equation}
These bounds prevent the policy from producing actions outside the valid workspace.

\textbf{Velocity constraints (VelCons)} bound one-step joint displacement:
\begin{gather}
    |\mathbf{q}_{t} - \mathbf{a}_{t}^{q}| \leq \mathbf{v}_{\max}, \\
    |\mathbf{a}_{t+i}^{q} - \mathbf{a}_{t+i+1}^{q}| \leq \mathbf{v}_{\max},
    \quad \forall i \in \{0,\dots,H-2\}.
\end{gather}
Unlike PosCons, VelCons depends on the current proprioceptive state $\mathbf{q}_t$, making the feasible set $\mathcal{C}(\mathbf{o}_t)$ state-dependent and time-varying. 

\paragraph{Backbones and baselines.}
We instantiate ActSafeGuard on two representative flow-matching policies: $\pi_{0.5}$~\cite{black2025pi05}, a pretrained VLA model, and Fast-WAM~\cite{yuan2026fastwam}, a world-action model that uses world modeling during training while retaining direct action prediction at inference time. We compare against three constraint-handling baselines: \emph{Projection}, an inference-time step-wise projection method; \emph{Truncation}, a post-hoc clipping method applied after unconstrained generation; and \emph{GaugeFlow}~\cite{li2026gauge}, a training-aware method based on an invertible gauge map. Detail introduction of baselines are in Appendix~B. GaugeFlow requires a static star-shaped feasible region, so it is only applicable to PosCons and is marked N/A under PosCons + VelCons.

\paragraph{Evaluation metrics.}
We report Success Rate (SR), Step Safe Rate (SSR), Maximum Mean Discrepancy (MMD), and Log Dimensionless Jerk (LDLJ)~\cite{balasubramanian2011robust}. SR measures task completion, SSR measures the percentage of executed control steps satisfying all constraints, MMD measures distributional deviation from expert action chunks, and LDLJ measures trajectory smoothness. MMD and LDLJ evaluate whether hard constraint enforcement damages the action-generation quality of the pretrained backbone. More details are in Appendix~C.

\begin{table}[t]
\centering
\small 
\begin{tabular}{lcccc}
\toprule
\multirow{2}{*}{Task}
& \multicolumn{2}{c}{PosCons}
& \multicolumn{2}{c}{PosCons + VelCons} \\
\cmidrule(lr){2-3} \cmidrule(lr){4-5}
& $\pi_{0.5}$ & Fast-WAM & $\pi_{0.5}$ & Fast-WAM \\
\midrule
\textit{lift pot}        & 83.59 & 15.12 & 82.85 & 13.68 \\
\textit{place shoe}      & 20.43 & 5.86  & 16.45 & 3.79 \\
\textit{hanging mug}     & 64.62 & 39.69 & 57.38 & 35.24 \\
\textit{place empty cup} & 37.04 & 31.27 & 35.69 & 27.33 \\
\midrule
\textbf{Mean SSR}        & \textbf{51.42} & \textbf{22.99} & \textbf{48.09} & \textbf{20.01} \\
\bottomrule
\end{tabular}
\caption{\textbf{Step Safety Rate (SSR, \%) of the unconstrained baseline with different backbones.}
Evaluated across four manipulation tasks under static position constraints (PosCons) and combined dynamic constraints (PosCons + VelCons).}
\label{tab:baseline_ssr_backbones}

\end{table}

\subsection{Why Are Hard Constraints Necessary?}
\label{sec:hard_constraint_necessity}

We first examine whether safety can be obtained simply by training on feasible demonstrations. For each task, the PosCons and VelCons bounds are constructed from the minimum and maximum statistics of the training data, so the training demonstrations lie inside the feasible set by design. We then fine-tune the unconstrained baselines on these data and evaluate them in in-distribution RoboTwin environments.

Table~\ref{tab:baseline_ssr_backbones} shows that feasible training data alone is insufficient for deterministic safety.  The mean SSR of the $\pi_{0.5}$ baseline is only $51.42\%$ under PosCons and $48.09\%$ under PosCons + VelCons, while Fast-WAM drops further to $22.99\%$ and $20.01\%$, respectively. These violations occur despite using in-distribution evaluation settings, indicating that neural fitting error, flow sampling error, and distribution shift can push generated chunks outside the empirical feasible envelope.

Therefore, safety cannot rely solely on data filtering or average-case regularization; it requires an explicit mechanism that guarantees per-step feasibility under the provided constraints.

\subsection{Performance Preservation (RQ1)}
\label{sec:performance_preservation}

\begin{figure}[t]
    \centering
    \begin{subfigure}[b]{\linewidth}
        \centering
        \includegraphics[width=\linewidth]{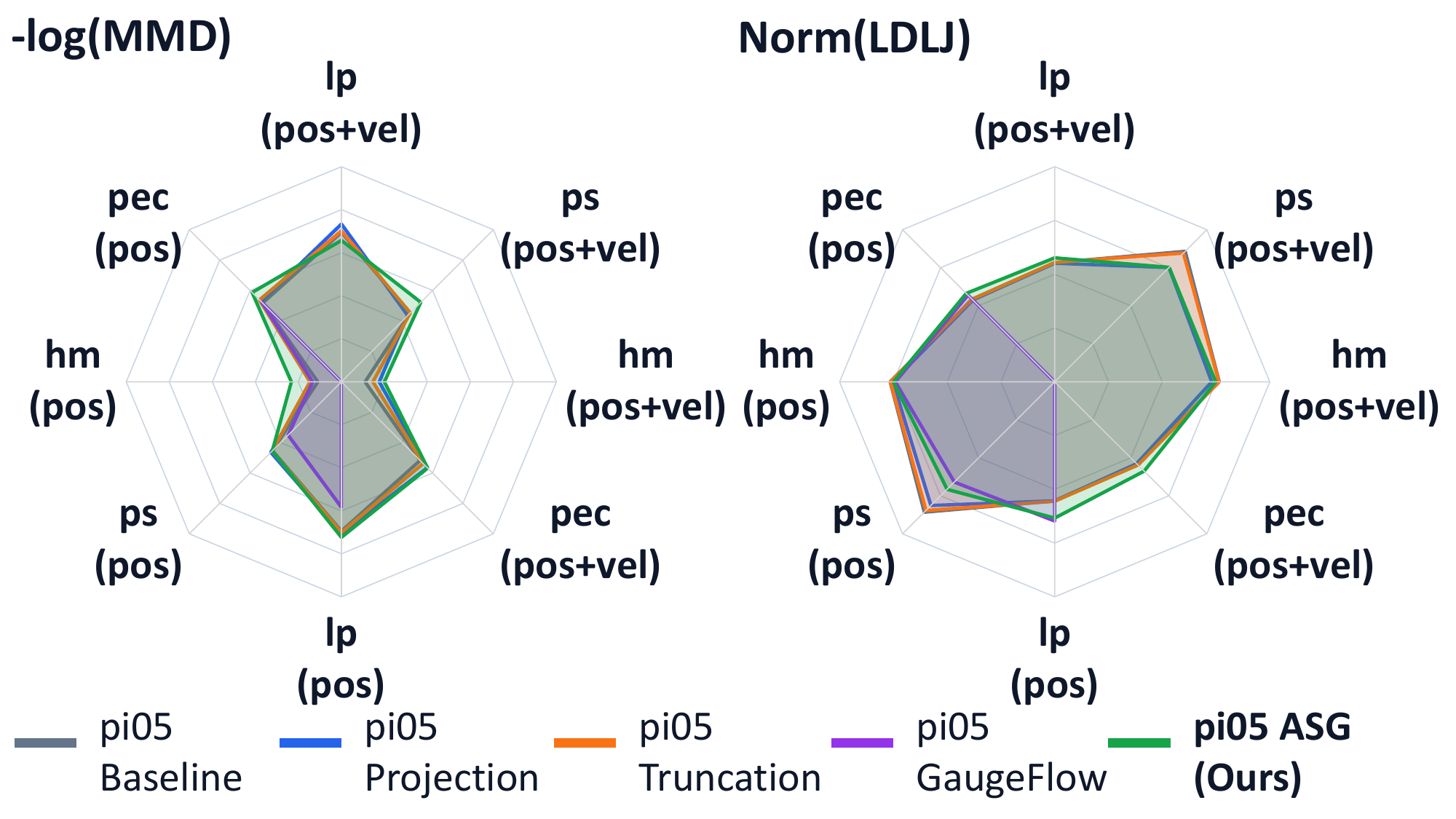}
        \caption{$\pi_{0.5}$ backbone}
        \label{fig:radar_pi05}
    \end{subfigure}
    \hfill
    \begin{subfigure}[b]{\linewidth}
        \centering
        \includegraphics[width=\linewidth]{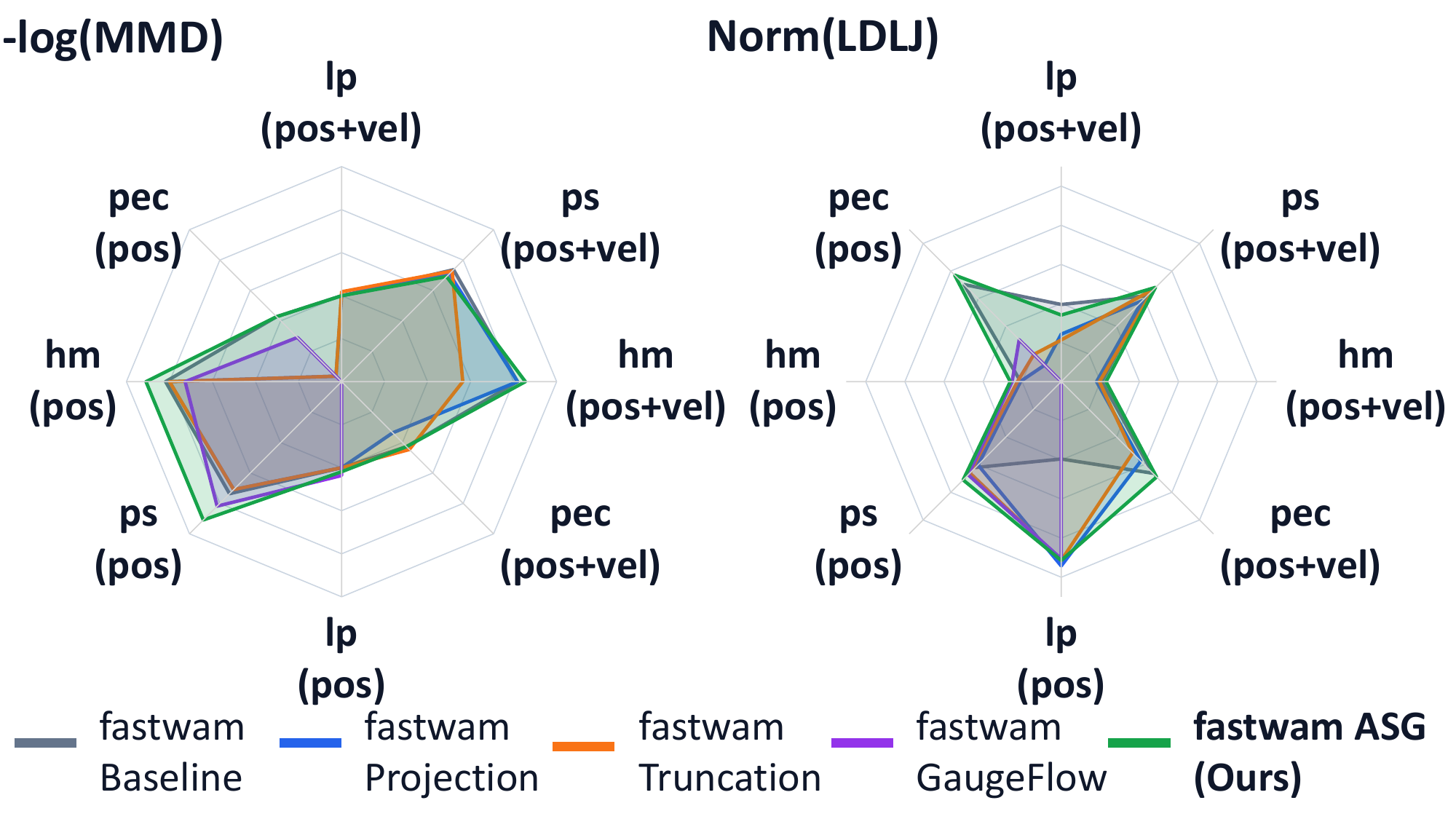}
        \caption{Fast-WAM backbone}
        \label{fig:radar_fastwam}
    \end{subfigure}
    
    \caption{\textbf{Trajectory quality and distribution alignment comparisons.} Radar charts of $-\log(\text{MMD})$ (higher is better) and normalized LDLJ (mean/std normalization, higher is better) across manipulation tasks: \textit{lift pot} (\textit{lp}), \textit{place shoe} (\textit{ps}), \textit{hanging mug} (\textit{hm}), and \textit{place empty cup} (\textit{pec}), under static position constraints (\textbf{pos}) and dynamic position + velocity constraints (\textbf{pos} + \textbf{vel}). Larger covered areas indicate superior performance in preserving original trajectory distributions and smoothness. Details in Appendix~E.}
    \label{fig:radar_overall}
\end{figure}

Table~\ref{tab:main_results_combined_highlight} evaluates task success rates under hard constraints across various scenarios. Under guaranteed safety, ActSafeGuard achieves the best success rate among constrained methods in most tasks, matching or even outperforming the unconstrained baseline in certain cases. 
The trajectory-quality results in Figure~\ref{fig:radar_overall} further show that ActSafeGuard preserves the pretrained action distribution and smoothness better than post-hoc correction methods.


Compared with inference-based methods, ActSafeGuard eliminates training--inference mismatch regarding constraint enforcement. Furthermore, its implicit oblique-projection gradients during training enable flexible boundary-sliding corrections, yielding lower fitting errors and smoother trajectories than simple truncation or projection. 

Compared with GaugeFlow, ActSafeGuard also provides a more conservative modification to the pretrained action head. GaugeFlow maps the original action space into a unit ball and trains the flow in the transformed coordinate system. This transformation can disrupt the vector-field priors already learned by the pretrained backbone. ActSafeGuard instead operates directly on the original flow update and only modifies the step when the nominal update would cross a boundary. This minimal intervention better preserves the pretrained capabilities.

\subsection{Generalizability to Complex Constraints (RQ2)}
\label{sec:complex_constraints}

We next evaluate whether ActSafeGuard scales beyond static box constraints. Under PosCons + VelCons, the feasible region can be written as a high-dimensional polytope $\mathbf{A}(\mathbf{o}_t)\mathbf{x} \leq \mathbf{b}(\mathbf{o}_t)$ whose facets depend on the current robot state and on adjacent actions within the generated chunk. This setting is substantially harder than static PosCons because the feasible set changes at every control step and couples multiple future actions through velocity bounds.

As shown in Table~\ref{tab:main_results_combined_highlight} and Figure~\ref{fig:radar_overall}, ActSafeGuard remains effective in this dynamic regime. ActSafeGuard obtains the highest mean SR under PosCons + VelCons, outperforming other baselines while maintaining deterministic safety.

This result highlights an important practical advantage of the ray-scaling formulation. Methods based on a fixed gauge map or other static coordinate transformations are difficult to apply when the feasible set changes with $\mathbf{o}_t$; this is why GaugeFlow is not applicable to the PosCons + VelCons regime. CBF-style inference methods also become challenging for high-dimensional action chunks with many coupled linear inequalities. In contrast, ActSafeGuard only requires computing ray-boundary intersections with the current feasible polytope. As a result, it can naturally support both static and dynamic constraints without changing the pretrained backbone or solving an iterative optimization problem during each sampling step.

\subsection{Ablation on Differentiability (RQ3)}
We isolate the contribution of differentiability by applying a stop-gradient operation to the safety scaling factor $s_{\phi}$ during fine-tuning. This variant, denoted ActSafeGuard (w/ sg), keeps the ray-scaling operator active during inference, so it still enforces $100\%$ SSR. However, the action head no longer receives boundary-aware gradients through $s_{\phi}$ during training. The comparison is shown in Table~\ref{tab:ablation_differentiability}.

The results demonstrate that deterministic inference-time feasibility is not sufficient for learning a useful constrained policy. On \emph{place shoe} under PosCons + VelCons, removing gradients through $s_{\phi}$ causes SR to collapse from $98.0\%$ to $3.0\%$. This large gap indicates that dynamic constraints require the policy to learn how its action chunks interact with moving feasible boundaries.

\begin{table}[t]
\centering
\small
\setlength{\tabcolsep}{5pt} 
\begin{tabular}{lccc}
\toprule
\textbf{Method} & \textbf{SR (\%)} $\uparrow$ & \textbf{LDLJ} $\uparrow$ & \textbf{MMD} $\downarrow$ \\ 
\midrule
\multicolumn{4} {l} {\textit{Task 1: place shoe (PosCons + VelCons)}} \\
$\pi_{0.5}$ + ActSafeGuard  & \textbf{98.0} & \textbf{-17.171} & \textbf{0.00242} \\ 
$\pi_{0.5}$ + ActSafeGuard (w/ sg) & 3.0 & -24.611 & 0.03008 \\ 
\midrule
\multicolumn{4} {l} {\textit{Task 2: place empty cup (PosCons)}} \\
$\pi_{0.5}$ + ActSafeGuard  & \textbf{97.0} & \textbf{-16.197} & \textbf{0.00960} \\ 
$\pi_{0.5}$ + ActSafeGuard (w/ sg) & 95.0 & -16.991 & 0.01171 \\ 
\bottomrule
\end{tabular}
\caption{\textbf{Ablation study on gradient differentiability.} Evaluation is conducted across different tasks and constraint configurations. ``w/ sg'' denotes applying a stop-gradient operation on the safety scaling factor during training. Note that the step safety rate (SSR) is omitted from the table as both variants deterministically achieve 100\% SSR across all evaluation episodes.}
\label{tab:ablation_differentiability}
\end{table}

\subsection{Real-Robot Deployment}
\label{sec:real_robot}

\begin{figure}[htbp]
    \centering
    \begin{subfigure}[b]{\linewidth}
        \centering
        \includegraphics[width=\linewidth]{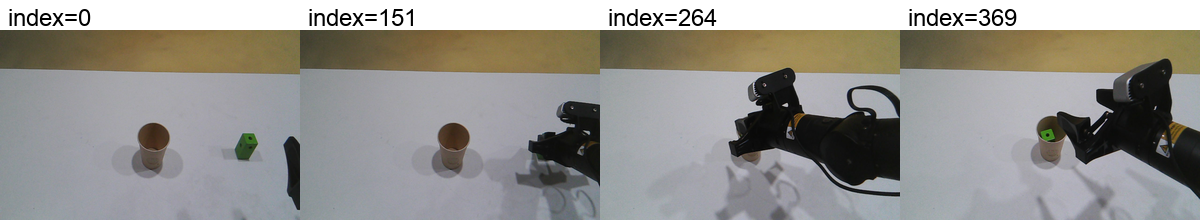}
        \caption{Snapshots on pick green cube task.}
        \label{fig:pick_green_cube}
    \end{subfigure}
    \hfill 
    \begin{subfigure}[b]{\linewidth}
        \centering
        \includegraphics[width=\linewidth]{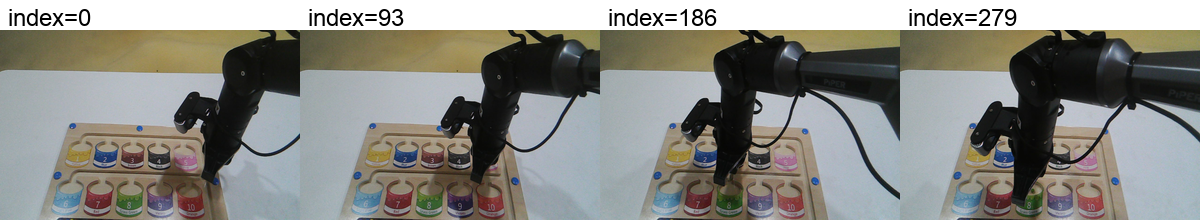}
        \caption{Snapshots on guide the ball task.}
        \label{fig:maze}
    \end{subfigure}
    \hfill
    \begin{subfigure}[b]{\linewidth}
        \centering
        \includegraphics[width=\linewidth]{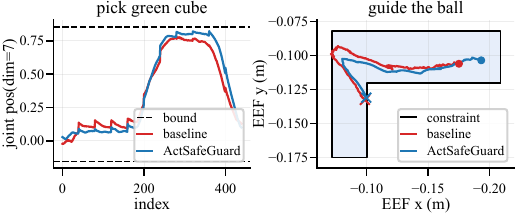}
        \caption{Action trajectories and constraints in two tasks.}
        \label{fig:real_traj}
    \end{subfigure}
    
    \caption{\textbf{Real Robot Deployment Results.} ActSafeGuard strictly satisfies constraints while maintaining task performance.}
    \label{fig:combined_figure}
\end{figure}

To assess the practical deployment applicability of ActSafeGuard, we evaluate our approach on a physical AgileX ALOHA bimanual platform. We design two distinct real-world tasks (details are in Appendix~F):
\textbf{\textit{pick green cube}:} The robotic arm is required to pick up a green cube and place it into a paper cup. The policy outputs absolute joint positions, and safety constraints are imposed as joint boundaries.
\textbf{\textit{guide the ball}:} Using a magnetic wand attached to the gripper, the arm guides a green ball along a physical track into a designated container. The policy outputs End-Effector (EEF) Cartesian coordinates, with safety constraints specified as an L-shaped corridor to prevent the EEF from slipping off the track. We adopt $\pi_{0.5}$ as the backbone and fine-tune it using $50$ demonstration trajectories collected for each task. Figure~\ref{fig:combined_figure} illustrates real-world execution snapshots and the corresponding execution trajectories. 
 
As depicted in the trajectory plots (Figure~\ref{fig:combined_figure}c), ActSafeGuard strictly adheres to the prescribed boundaries across all steps. In contrast, the unconstrained baseline violates the corridor boundary in \textit{guide the ball}. We evaluate each method across $5$ rollouts per task. For \textit{pick green cube}, both the baseline and ActSafeGuard achieve $5/5$ successes. For the more constrained \textit{guide the ball} task, the baseline achieves $4/5$ successes, whereas ActSafeGuard succeeds in all $5/5$ trials. These empirical results show that ActSafeGuard can guarantee constraint satisfaction without compromising task capability. Notably, the feasible set in \textit{guide the ball} forms a non-convex L-shaped corridor, further suggesting the generality of ActSafeGuard: as long as the ray-boundary intersection can be computed, the same ray-scaling method can be extended beyond convex polytopes.

\section{Conclusion}
We present ActSafeGuard, a differentiable and training-aligned safeguard layer for flow-matching based VLA and WAM policies. By applying a parameter-free ray-scaling operator to each discrete flow update, ActSafeGuard enforces hard action feasibility at every generation step. Its local Jacobian induces boundary-sliding gradients, enabling the action head to adapt to feasible-set geometry rather than relying on post-hoc correction alone. Across simulation and real robot tasks, multiple constraint regimes, and two foundation backbones, ActSafeGuard achieved deterministic step safety while maintaining strong task success and action-generation quality. 
ActSafeGuard also has some limitations. Currently it assumes that the feasible action space can be represented by explicitly specified constraints for which ray-boundary intersections are tractable. Moreover, the constraint set is still manually specified from task knowledge or demonstration statistics. Automatically discovering, validating, and updating such constraints from perception or interaction data is an important direction for future work.


\appendix

\bibliography{aaai2027}

@misc{kim2024openvla,
  title = {{OpenVLA}: An Open-Source Vision-Language-Action Model},
  author = {Kim, Moo Jin and Pertsch, Karl and Karamcheti, Siddharth and Xiao, Ted and Balakrishna, Ashwin and Nair, Suraj and Rafailov, Rafael and Foster, Ethan and Lam, Grace and Sanketi, Pannag and Vuong, Quan and Kollar, Thomas and Burchfiel, Benjamin and Tedrake, Russ and Sadigh, Dorsa and Levine, Sergey and Liang, Percy and Finn, Chelsea},
  year = {2024},
  eprint = {2406.09246},
  archivePrefix = {arXiv},
  primaryClass = {cs.RO}
}

@misc{black2025pi05,
  title = {$\pi_{0.5}$: A Vision-Language-Action Model with Open-World Generalization},
  author = {Black, Kevin and Brown, Noah and Darpinian, James and Dhabalia, Karan and Driess, Danny and Esmail, Adnan and Equi, Michael and Finn, Chelsea and Fusai, Niccolo and Galliker, Manuel Y. and Ghosh, Dibya and Groom, Lachy and Hausman, Karol and Ichter, Brian and Jakubczak, Szymon and Jones, Tim and Ke, Liyiming and LeBlanc, Devin and Levine, Sergey and Li-Bell, Adrian and Mothukuri, Mohith and Nair, Suraj and Pertsch, Karl and Ren, Allen Z. and Shi, Lucy Xiaoyang and Smith, Laura and Springenberg, Jost Tobias and Stachowicz, Kyle and Tanner, James and Vuong, Quan and Walke, Homer and Walling, Anna and Wang, Haohuan and Yu, Lili and Zhilinsky, Ury},
  year = {2025},
  eprint = {2504.16054},
  archivePrefix = {arXiv},
  primaryClass = {cs.RO}
}

@misc{nvidia2025gr00tn1openfoundation,
      title={GR00T N1: An Open Foundation Model for Generalist Humanoid Robots}, 
      author={NVIDIA and : and Johan Bjorck and Fernando Castañeda and Nikita Cherniadev and Xingye Da and Runyu Ding and Linxi "Jim" Fan and Yu Fang and Dieter Fox and Fengyuan Hu and Spencer Huang and Joel Jang and Zhenyu Jiang and Jan Kautz and Kaushil Kundalia and Lawrence Lao and Zhiqi Li and Zongyu Lin and Kevin Lin and Guilin Liu and Edith Llontop and Loic Magne and Ajay Mandlekar and Avnish Narayan and Soroush Nasiriany and Scott Reed and You Liang Tan and Guanzhi Wang and Zu Wang and Jing Wang and Qi Wang and Jiannan Xiang and Yuqi Xie and Yinzhen Xu and Zhenjia Xu and Seonghyeon Ye and Zhiding Yu and Ao Zhang and Hao Zhang and Yizhou Zhao and Ruijie Zheng and Yuke Zhu},
      year={2025},
      eprint={2503.14734},
      archivePrefix={arXiv},
      primaryClass={cs.RO}
}

@misc{ye2026worldactionmodels,
  title = {World Action Models are Zero-shot Policies},
  author = {Ye, Seonghyeon and Ge, Yunhao and Zheng, Kaiyuan and Gao, Shenyuan and Yu, Sihyun and Kurian, George and Indupuru, Suneel and Tan, You Liang and Zhu, Chuning and Xiang, Jiannan and Malik, Ayaan and Lee, Kyungmin and Liang, William and Ranawaka, Nadun and Gu, Jiasheng and Xu, Yinzhen and Wang, Guanzhi and Hu, Fengyuan and Narayan, Avnish and Bjorck, Johan and Wang, Jing and Kim, Gwanghyun and Niu, Dantong and Zheng, Ruijie and Xie, Yuqi and Wu, Jimmy and Wang, Qi and Julian, Ryan and Xu, Danfei and Du, Yilun and Chebotar, Yevgen and Reed, Scott and Kautz, Jan and Zhu, Yuke and Fan, Linxi and Jang, Joel},
  year = {2026},
  eprint = {2602.15922},
  archivePrefix = {arXiv},
  primaryClass = {cs.RO}
}

@misc{liao2025genieenvisioner,
  title = {Genie Envisioner: A Unified World Foundation Platform for Robotic Manipulation},
  author = {Liao, Yue and Zhou, Pengfei and Huang, Siyuan and Yang, Donglin and Chen, Shengcong and Jiang, Yuxin and Hu, Yue and Cai, Jingbin and Liu, Si and Luo, Jianlan and Chen, Liliang and Yan, Shuicheng and Yao, Maoqing and Ren, Guanghui},
  year = {2025},
  eprint = {2508.05635},
  archivePrefix = {arXiv},
  primaryClass = {cs.RO}
}

@misc{yuan2026fastwam,
  title = {{Fast-WAM}: Do World Action Models Need Test-time Future Imagination?},
  author = {Yuan, Tianyuan and Dong, Zibin and Liu, Yicheng and Zhao, Hang},
  year = {2026},
  eprint = {2603.16666},
  archivePrefix = {arXiv},
  primaryClass = {cs.RO}
}

@article{zhang2026safevla,
  title={Safevla: Towards safety alignment of vision-language-action model via constrained learning},
  author={Zhang, Borong and Zhang, Yuhao and Ji, Jiaming and Lei, Yingshan and Dai, Juntao and Chen, Yuanpei and Yang, Yaodong},
  journal={Advances in Neural Information Processing Systems},
  volume={38},
  pages={153335--153373},
  year={2026}
}

@misc{tang2026safedojo,
  title = {{SafeDojo}: Safe Reinforcement Learning for {VLA} via Interactive World Model},
  author = {Tang, Kai and Jia, Peidong and Chu, Zhong and Wu, Jixian and Ma, Rui and Cao, Jiajun and Zhao, Fangyuan and Chen, Sixiang and Guo, Yichen and Chi, Xiaowei and Fan, Chun-Kai and Zhang, Kevin and Xu, Jinchang and Yang, Fubing and Mi, Weishi and Ju, Xiaozhu and Tang, Jian and Zhang, Shanghang},
  year = {2026},
  eprint = {2606.20698},
  archivePrefix = {arXiv},
  primaryClass = {cs.RO}
}

@misc{wei2026can,
      title={Can Explicit Physical Feasibility Benefit VLA Learning? An Empirical Study}, 
      author={Yubai Wei and Chen Wu and Hashem Haghbayan},
      year={2026},
      eprint={2604.17896},
      archivePrefix={arXiv},
      primaryClass={cs.LG}
}

@misc{hu2025vlsa,
  title = {{VLSA}: Vision-Language-Action Models with Plug-and-Play Safety Constraint Layer},
  author = {Hu, Songqiao and Liu, Zeyi and Liu, Shuang and Cen, Jun and Meng, Zihan and Wang, Shihefeng and Li, Xiang and He, Xiao},
  year = {2025},
  eprint = {2512.11891},
  archivePrefix = {arXiv},
  primaryClass = {cs.RO}
}

@misc{english2026neurosymbolic,
  title = {Neuro-Symbolic Safety Guidance for Vision-Language-Action Models via Constrained Flow Matching},
  author = {English, William and Zheng, Hao and Ewetz, Rickard},
  year = {2026},
  eprint = {2607.01378},
  archivePrefix = {arXiv},
  primaryClass = {cs.RO}
}

@misc{beaudin2026anybodyguard,
  title = {Any-Body Guard: Universal Safeguarding for Manipulation Policies via Action Masking},
  author = {Beaudin, Alex and Krasowski, Hanna and Nagpal, Kartik and Seshia, Sanjit A. and Arcak, Murat and Mehr, Negar},
  year = {2026},
  eprint = {2606.22278},
  archivePrefix = {arXiv},
  primaryClass = {cs.RO}
}

@article{chandra2026physvla,
  title={PhysVLA: Towards Physically-Grounded VLA for Embodied Robotic Manipulation},
  author={Chandra, Namai and Damodaran, Shriram and Wang, Lin},
  journal={arXiv preprint arXiv:2606.13886},
  year={2026}
}

@misc{li2026embodiedaisafety,
  title = {Safety in Embodied {AI}: A Survey of Risks, Attacks, and Defenses},
  author = {Li, Xiao and Zheng, Xiang and Gao, Yifeng and Xia, Xinyu and Wang, Yixu and Wang, Xin and Sun, Ye and Zhao, Yunhan and Wen, Ming and Li, Jiayu and Chen, Zixing and Gong, Xun and Liu, Yi and Li, Yige and Wu, Yutao and Wang, Cong and Sun, Jun and Cao, Yixin and Chen, Zhineng and Chen, Jingjing and Gui, Tao and Zhang, Qi and Wu, Zuxuan and Qiu, Xipeng and Huang, Xuanjing and Zhang, Tiehua and Wei, Zhipeng and Wang, Kun and Li, Xinfeng and Huang, Hanxun and Erfani, Sarah and Bailey, James and Wang, Jianping and Xiao, Chaowei and He, Ran and Li, Bo and Ma, Xingjun and Jiang, Yu-Gang},
  year = {2026},
  eprint = {2605.02900},
  archivePrefix = {arXiv},
  primaryClass = {cs.CR}
}

@misc{li2026vlasafety,
  title = {Vision-Language-Action Safety: Threats, Challenges, Evaluations, and Mechanisms},
  author = {Li, Qi and Yin, Bo and Huang, Weiqi and Liu, Ruhao and Zou, Bojun and Yu, Runpeng and Ye, Jingwen and Yu, Weihao and Wang, Xinchao},
  year = {2026},
  eprint = {2604.23775},
  archivePrefix = {arXiv},
  primaryClass = {cs.RO}
}

@misc{kim2026safeembodied,
  title = {Safe Embodied {AI} for Long-horizon Tasks: A Cross-layer Analysis of Robotic Manipulation},
  author = {Kim, Dabin and Park, Daemin and Lee, Sangyub and Kim, Jinsik and Oh, Yeongtak and Shin, Jongho and Yoon, Sungroh},
  year = {2026},
  eprint = {2606.05660},
  archivePrefix = {arXiv},
  primaryClass = {cs.RO}
}

@misc{kim2026modularguardrails,
  title = {Modular Safety Guardrails Are Necessary for Foundation-Model-Enabled Robots in the Real World},
  author = {Kim, Joonkyung and Chen, Wenxi and Soleymanzadeh, Davood and Ding, Yi and Gao, Xiangbo and Tu, Zhengzhong and Zhang, Ruqi and Fei, Fan and Veer, Sushant and Lyu, Yiwei and Zheng, Minghui and Gu, Yan},
  year = {2026},
  eprint = {2602.04056},
  archivePrefix = {arXiv},
  primaryClass = {cs.RO}
}

@inproceedings{
lipman2023flow,
title={Flow Matching for Generative Modeling},
author={Yaron Lipman and Ricky T. Q. Chen and Heli Ben-Hamu and Maximilian Nickel and Matthew Le},
booktitle={The Eleventh International Conference on Learning Representations },
year={2023}
}

@inproceedings{ma2026polyflow,
    title={PolyFlow: Safe and Efficient Polytope-Constrained Flow Matching with Constraint Embedding and Projection-free Update},
    author={Jianming Ma and Qiyue Yang and Yang Zhang and Liyun Yan and Zhanxiang Cao and Yazhou Zhang and Yue Gao},
    booktitle={Forty-third International Conference on Machine Learning},
    year={2026}
}

@inproceedings{li2026gauge,
  title={Gauge Flow Matching: Efficient Constrained Generative Modeling over General Convex Set and Beyond},
  author={Li, Xinpeng and Liang, Enming and Chen, Minghua},
  booktitle={The Fourteenth International Conference on Learning Representations},
  year={2026}
}

@article{balasubramanian2011robust,
  title={A robust and sensitive metric for quantifying movement smoothness},
  author={Balasubramanian, Sivakumar and Melendez-Calderon, Alejandro and Burdet, Etienne},
  journal={IEEE transactions on biomedical engineering},
  volume={59},
  number={8},
  pages={2126--2136},
  year={2011},
  publisher={IEEE}
}

@misc{chen2025robotwin,
      title={RoboTwin 2.0: A Scalable Data Generator and Benchmark with Strong Domain Randomization for Robust Bimanual Robotic Manipulation}, 
      author={Tianxing Chen and Zanxin Chen and Baijun Chen and Zijian Cai and Yibin Liu and Zixuan Li and Qiwei Liang and Xianliang Lin and Yiheng Ge and Zhenyu Gu and Weiliang Deng and Yubin Guo and Tian Nian and Xuanbing Xie and Qiangyu Chen and Kailun Su and Tianling Xu and Guodong Liu and Mengkang Hu and Huan-ang Gao and Kaixuan Wang and Zhixuan Liang and Yusen Qin and Xiaokang Yang and Ping Luo and Yao Mu},
      year={2025},
      eprint={2506.18088},
      archivePrefix={arXiv},
      primaryClass={cs.RO}
}

\newpage


\section{Appendix}

\subsection{Appendix A: Parameterizations of Constraints}
\label{sec:appendix_a}

We elaborate on the concrete parameterizations of the two constraint families utilized in our main experiments: Position Constraints (\textbf{PosCons}) and Velocity Constraints (\textbf{VelCons}). Both constraint formulations operate on the $12$ primary arm-joint dimensions (corresponding to indices $\{0,1,2,3,4,5,7,8,9,10,11,12\}$ within the $14$-dimensional ALOHA-Agilex action vector), while leaving the two gripper dimensions (indices $6$ and $13$) unconstrained. We derive per-task numerical bounds from two complementary data sources:
\begin{itemize}
    \item \textbf{Demonstration Statistics:} We empirically compute the maximum and minimum values across each action dimension in the training dataset to establish the positional boundaries ($\mathbf{q}_{\min}, \mathbf{q}_{\max}$). Additionally, the maximum absolute difference between consecutive timesteps across the trajectories defines the maximum permissible per-step velocity limit ($\mathbf{v}_{\max}$).
    \item \textbf{Hardware Specifications:} The derived \textbf{PosCons} bounding boxes are cross-checked against the official ALOHA-Agilex bimanual manipulator documentation. Any statistical bound exceeding the physical joint envelope is strictly clamped back to the manufacturer's recommended soft limits.
\end{itemize}

The detailed numerical bounds for \textbf{PosCons} and \textbf{VelCons} across all four manipulation tasks are summarized in Table~\ref{tab:appendix_poscons} and Table~\ref{tab:appendix_velcons}, respectively.

\begin{table}[htbp]
\centering
\small
\setlength{\tabcolsep}{3pt}

\begin{subtable}{\linewidth}
\centering
\begin{tabular}{ll}
\toprule
\textbf{Task} & Lower Bounds $\mathbf{q}_{\min}$ (rad) \\
\midrule
\textit{lift pot} & [-0.331, 0.000, 0.000, -1.949, -0.000, -0.998, \\
                  & \phantom{[}-0.087, 0.000, 0.000, -2.003, -1.046, -0.507] \\
\midrule
\textit{place shoe} & [-0.846, 0.000, 0.000, -1.924, -0.216, -2.405, \\
                    & \phantom{[}-0.059, 0.000, 0.000, -1.924, -0.158, -3.990] \\
\midrule
\textit{hanging mug} & [-1.088, -0.000, -0.000, -1.951, -0.134, -1.194, \\
                     & \phantom{[}-3.310, -0.784, 0.000, -1.604, -1.997, -3.893] \\
\midrule
\textit{place empty cup} & [-0.918, 0.000, 0.000, -1.781, -0.007, -0.894, \\
                         & \phantom{[}-0.151, 0.000, 0.000, -1.892, -0.095, -0.102] \\
\bottomrule
\end{tabular}
\caption{Lower bounds ($\mathbf{q}_{\min}$ in radians)}
\label{tab:appendix_poscons_qmin}
\end{subtable}

\vspace{2mm} 

\begin{subtable}{\linewidth}
\centering
\begin{tabular}{ll}
\toprule
\textbf{Task} & Upper Bounds $\mathbf{q}_{\max}$ (rad) \\
\midrule
\textit{lift pot} & [0.043, 2.779, 2.900, 0.366, 1.168, 0.627, \\
                  & \phantom{[}0.364, 2.790, 2.930, 0.387, 0.004, 6.614] \\
\midrule
\textit{place shoe} & [0.000, 2.870, 3.234, 0.000, 0.349, 1.872, \\
                    & \phantom{[}6.992, 2.845, 3.200, 0.000, 0.239, 2.356] \\
\midrule
\textit{hanging mug} & [0.000, 2.686, 3.219, 0.000, 0.111, 1.183, \\
                     & \phantom{[}4.214, 2.602, 5.129, 1.958, 1.126, 3.152] \\
\midrule
\textit{place empty cup} & [0.127, 2.684, 2.903, 0.000, 0.096, 0.149, \\
                         & \phantom{[}0.881, 3.009, 3.372, 0.000, 0.101, 0.899] \\
\bottomrule
\end{tabular}
\caption{Upper bounds ($\mathbf{q}_{\max}$ in radians)}
\label{tab:appendix_poscons_qmax}
\end{subtable}

\caption{\textbf{Per-task Position Constraint (PosCons) bounds.}}
\label{tab:appendix_poscons}
\end{table}

\begin{table}[htbp]
\centering
\small
\setlength{\tabcolsep}{4pt}
\begin{tabular}{ll}
\toprule
\textbf{Task} & $\mathbf{v}_{\max}$ per joint (rad) \\
\midrule
\textit{lift pot} & [0.015, 0.126, 0.104, 0.078, 0.070, 0.041, \\
                  & \phantom{[}0.013, 0.133, 0.110, 0.077, 0.056, 0.176] \\
\midrule
\textit{place shoe} & [0.067, 0.156, 0.331, 0.129, 0.037, 0.167, \\
                    & \phantom{[}3.982, 0.134, 0.296, 0.104, 0.020, 0.165] \\
\midrule
\textit{hanging mug} & [0.074, 0.143, 0.316, 0.126, 0.016, 0.074, \\
                     & \phantom{[}0.146, 0.523, 0.164, 0.122, 0.113, 0.136] \\
\midrule
\textit{place empty cup} & [0.083, 0.127, 0.124, 0.094, 0.013, 0.082, \\
                         & \phantom{[}0.080, 0.127, 0.123, 0.094, 0.012, 0.081] \\
\bottomrule
\end{tabular}
\caption{\textbf{Per-task Velocity Constraint (VelCons) bounds.} $\mathbf{v}_{\max}$ is the maximum permissible joint displacement per control step, computed from the demonstration statistics.}
\label{tab:appendix_velcons}
\end{table}


\paragraph{Action-Space Decomposition and Ray-Shooting Parameterization}
At each policy query, the action head predicts an action chunk with shape $[H, D]$, where $H$ denotes the chunk horizon and $D=14$ is the dimensionality of the ALOHA-Agilex action space. We decompose the predicted action chunk into a constrained arm-joint component of shape $[H, D_c]$ and an unconstrained gripper component of shape $[H, D_u]$, where $D_c=12$ and $D_u=2$. ActSafeGuard is applied only to the constrained arm-joint dimensions, while the unconstrained gripper dimensions are directly passed through without modification.

For different constraint types, we instantiate the ray-shooting operation in different spaces:
\begin{itemize}
    \item For tasks with \textbf{PosCons}, ActSafeGuard performs ray-shooting independently for each horizon step. In this case, the feasible set is defined over the $D_c=12$ constrained joint dimensions.
    \item For tasks with \textbf{PosCons} + \textbf{VelCons}, we adopt a joint-wise parallel parameterization. Specifically, for each constrained joint dimension, we collect its trajectory over the horizon as an $H$-dimensional vector and perform ray-shooting in this temporal space. This formulation is particularly suitable for constraints that couple actions across time, such as velocity limits, while also reducing the dimensionality of each ray-shooting problem.
\end{itemize}

\subsection{Appendix B: Implementation Details of Backbones and Baselines}
\label{sec:appendix_b}

\paragraph{Backbones.}
We evaluate ActSafeGuard on two representative flow-matching policies with substantially different backbone designs.

\textbf{$\pi_{0.5}$}~\cite{black2025pi05} is a classical Vision-Language-Action (VLA) model consisting of a pretrained multimodal VLM encoder and a flow-matching action expert. The VLM branch consumes multi-view RGB images and the language instruction, projects them into a shared token stream, and passes the fused representation to a Transformer-based action expert. The action expert regresses the flow-matching velocity field over a chunk of $H = 50$ future actions. In our experiments, we use the publicly released $\pi_{0.5}$ backbone and fine-tune all parameters (both VLM backbone and action expert) together with the ActSafeGuard operator.

\textbf{Fast-WAM}~\cite{yuan2026fastwam} is a World-Action Model that follows a Mixture-of-Transformers dual-branch design. A video-DiT branch models future frames from an input image, an action-DiT branch predicts actions, and both are trained jointly with mixed attention that ties the two modalities during training. At inference time only the action-DiT branch is invoked, which decouples the acted robot control loop from the expensive video-imagination path and enables real-time execution. The Wan2.2-TI2V-5B backbone is used for the video branch; the action-DiT is a 30-layer, 1024-hidden Transformer conditioned on proprioception, text, and image tokens. The whole model has around 6.02B parameters. We use the discrete-time flow-matching schedule with $N=20$ sampling steps for the action branch when combined with ActSafeGuard, and inherit the default video-branch schedule for training-only video co-supervision.

\paragraph{Baselines.}
We compare ActSafeGuard against three representative constraint-handling methods, spanning both inference-only and joint training-inference paradigms.

\begin{itemize}
    \item \textbf{Projection} (\emph{Inference-only}) implements a step-by-step projection during the flow-matching sampling process. At each inference step, if the intermediate sample violates the constraint polytope $\mathcal{C}(\mathbf{o}_t)$, it is projected onto the boundary of the feasible polytope. The training objective of the underlying backbone is unchanged, so the policy is optimized in an unconstrained space and executes in a constrained one.
    \item \textbf{Truncation (Trunc)} (\emph{Inference-only}) is a naive post-hoc correction. The unconstrained backbone generates a complete action chunk, and a single-shot box (and, when VelCons is present, per-step velocity) clipping is applied to $\mathbf{x}_N$ at the end of inference. For the velocity-constrained setting we additionally use a three-stage post-processor: (i) state-to-first clipping of $\mathbf{a}_t^q$ around the current proprioception $\mathbf{q}_t$; (ii) global box clipping across the horizon; and (iii) adjacent-step velocity clipping propagated left-to-right. This is the simplest but also the most distribution-distorting form of enforcement.
    \item \textbf{GaugeFlow}~\cite{li2026gauge} (\emph{Training \& Inference}) constructs an invertible coordinate transformation --- the gauge map --- from a star-shaped feasible polytope onto the unit ball. Flow matching is trained and sampled entirely in the transformed ball, then mapped back to physical space via the inverse gauge. GaugeFlow provides deterministic per-step feasibility, but the invertible construction currently only applies to static, star-shaped constraint sets; as such, we only report GaugeFlow numbers under \textbf{PosCons} and leave the corresponding \textbf{PosCons + VelCons} entries as \textbf{N/A}.
\end{itemize}

For all baselines we use the same backbone weights and the same training data as ActSafeGuard.

\subsection{Appendix C: Details of Evaluation Metrics}
\label{sec:appendix_c}

We now describe how the two distribution- and smoothness-level metrics used in the experiment section, \textbf{MMD} and \textbf{LDLJ}, are computed.

\paragraph{Maximum Mean Discrepancy (MMD).}
MMD measures the discrepancy between the generated action-chunk distribution and the expert action-chunk distribution from the training dataset. For each method and task, we sample $N$ ground-truth chunks $\{\mathbf{x}^{gt}_i\}_{i=1}^{N}$ from the training set and generate the corresponding predicted chunks $\{\mathbf{x}^{pr}_i\}_{i=1}^{N}$ using the evaluated policy. In our implementation, each action chunk has shape $H \times D$ with $D=14$, including both arm-joint and gripper dimensions. Each chunk is flattened into a vector in $\mathbb{R}^{H D}$ before computing distributional distances.

We use the biased empirical estimate of the squared MMD with an RBF kernel:
\begin{equation}
\begin{gathered}
\mathrm{MMD}^2(X, Y) =
     \frac{1}{N^2} \sum_{i=1}^{N}\sum_{j=1}^{N} k(x_i, x_j) \\
      + \frac{1}{N^2} \sum_{i=1}^{N}\sum_{j=1}^{N} k(y_i, y_j) 
     - \frac{2}{N^2} \sum_{i=1}^{N}\sum_{j=1}^{N} k(x_i, y_j),
\end{gathered}
\end{equation}
where $X=\{x_i\}_{i=1}^{N}$ denotes the flattened expert chunks and
$Y=\{y_i\}_{i=1}^{N}$ denotes the flattened generated chunks. The RBF kernel is
\begin{equation}
    k(u, v) = \exp\left(-\frac{\|u-v\|_2^2}{2\sigma^2}\right).
\end{equation}

The bandwidth $\sigma$ is selected using the median heuristic computed only from pairwise distances among the expert samples $X$:
\begin{equation}
    \sigma = \sqrt{
        \operatorname{median}
        \left(
        \left\{ \|x_i - x_j\|_2^2 \;:\; i \neq j \right\}
        \right)
    }.
\end{equation}
This choice keeps the kernel bandwidth fixed across different methods evaluated on the same task and sampled with the same random seed, avoiding method-dependent bandwidth estimates.

The reported metric in our tables is
\begin{equation}
    \mathrm{MMD}_{\mathrm{RBF}} = \sqrt{\mathrm{MMD}^2(X,Y)} ,
\end{equation}
Therefore, lower values indicate a closer match to the expert action distribution. In our experiments, we use $N=512$ sampled chunks per method-task pair with a fixed random seed.

\paragraph{Log Dimensionless Jerk (LDLJ).}
LDLJ~\cite{balasubramanian2011robust} is a scale-invariant smoothness metric that has become a standard descriptor of robot trajectory quality. Given an executed joint-position trajectory $\mathbf{q}_{0:T-1}$ (with 12 constrained joints and unit control-step $\Delta t = 1$), we approximate velocity by first-order differences and jerk by third-order differences:
\begin{equation}
    \dot{\mathbf{q}}_t \approx \frac{\mathbf{q}_{t+1} - \mathbf{q}_t}{\Delta t}, 
    \qquad
    \dddot{\mathbf{q}}_t \approx \frac{\Delta^3 \mathbf{q}_t}{\Delta t^3}.
\end{equation}
Let $T_{\mathrm{dur}} = (T-1)\Delta t$ denote the total trajectory duration, and let
$v_{\mathrm{peak}} = \max_t \|\dot{\mathbf{q}}_t\|_2$ be the peak joint-space speed. The dimensionless jerk cost is computed as
\begin{equation}
    \mathrm{DJ}
    =
    \frac{T_{\mathrm{dur}}^3}{v_{\mathrm{peak}}^2 + \varepsilon}
    \sum_t \|\dddot{\mathbf{q}}_t\|_2^2 \Delta t,
\end{equation}
and LDLJ is defined as
\begin{equation}
    \mathrm{LDLJ} = -\log\left(\mathrm{DJ} + \varepsilon\right),
\end{equation}
where $\varepsilon = 10^{-12}$ is used for numerical stability. Higher LDLJ values indicate smoother trajectories.

We compute LDLJ independently for each evaluation episode and report the mean across all episodes in a run. 

\subsection{Appendix D: Training and Evaluation Details}
\label{sec:appendix_d}

For each task, we use the data generation module in RoboTwin2 to collect $300$ successful trajectories as the training dataset. During evaluation, we use seeds that are disjoint from those used for data generation and evaluate each task over $100$ episodes.

In all experiments, we use identical training and evaluation configurations for the baseline methods and ActSafeGuard. When ActSafeGuard is applied, the discrete-time flow matching loss is computed only over the $12$ constrained action dimensions, while the loss terms for the remaining unconstrained dimensions are kept unchanged.

Table~\ref{tab:appendix_fastwam_details} lists the training and evaluation configurations used for Fast-WAM-based methods. Table~\ref{tab:appendix_pi05_details} lists the training and evaluation configurations used for $\pi_{0.5}$-based methods.

\begin{table}[t]
\centering
\setlength{\tabcolsep}{6pt}

\begin{subtable}{\linewidth}
\centering
\begin{tabular}{l c}
    \toprule
    \textbf{Parameter} & \textbf{Value} \\
    \midrule
    Action horizon $H$ & 32 \\
    Model precision & bfloat16 \\
    Optimizer & AdamW \\
    Weight decay & $1.0\times 10^{-2}$ \\
    Max grad norm & 1.0 \\
    LR scheduler & cosine \\
    Learning rate & $1.0\times 10^{-4}$ \\
    $\lambda_{\text{action}}$ & 1.0 \\
    Batch size & 4 \\
    Gradient accumulation steps & 1 \\
    Number of training steps & $2.0\times 10^{5}$ \\
    ActSafeGuard discrete steps $N$ & 20 \\
    Training platform & H200 $\times$ 8 \\
    \bottomrule
\end{tabular}
\caption{Training details}
\label{tab:appendix_fastwam_training}
\end{subtable}

\vspace{2mm}

\begin{subtable}{\linewidth}
\centering
\begin{tabular}{l c}
    \toprule
    \textbf{Parameter} & \textbf{Value} \\
    \midrule
    Execution horizon (replan steps) & 24 \\
    Number of sampling steps & 20 \\
    Sigma shift (action) & 1.0 \\
    Text CFG scale & 1.0 \\
    Evaluation platform & H200 $\times$ 1 \\
    Number of evaluation episodes & 100 \\
    \bottomrule
\end{tabular}
\caption{Evaluation details}
\label{tab:appendix_fastwam_evaluation}
\end{subtable}

\caption{\textbf{Hyperparameter details for Fast-WAM based methods.}}
\label{tab:appendix_fastwam_details}
\end{table}

\begin{table}[t]
\centering
\setlength{\tabcolsep}{6pt}

\begin{subtable}{\linewidth}
\centering
\begin{tabular}{l c}
    \toprule
    \textbf{Parameter} & \textbf{Value} \\
    \midrule
    Internal action dimension & 32 \\
    Executed action dimension & 14 \\
    Action horizon $H$ & 50 \\
    Model precision & bfloat16 \\
    Optimizer & AdamW \\
    Weight decay & $1.0\times 10^{-10}$ \\
    Max gradient norm & 1.0 \\
    LR scheduler & warmup cosine decay \\
    Warmup steps & 1,000 \\
    Peak learning rate & $2.5\times 10^{-5}$ \\
    Final learning rate & $2.5\times 10^{-6}$ \\
    LR decay steps & $3.0\times 10^{-4}$ \\
    batch size & 32 \\
    Gradient accumulation steps & 1 \\
    Number of training steps & $5.0\times 10^{4}$ \\
    ActSafeGuard discrete steps $N$ & 10 \\
    Training platform & H100 $\times$ 4 \\
    \bottomrule
\end{tabular}
\caption{Training details}
\label{tab:appendix_pi05_training}
\end{subtable}

\vspace{2mm}

\begin{subtable}{\linewidth}
\centering
\begin{tabular}{l c}
    \toprule
    \textbf{Parameter} & \textbf{Value} \\
    \midrule
    Execution horizon (replan steps) & 20 \\
    Number of sampling steps & 10 \\
    Evaluation platform & H100 $\times$ 1 \\
    Number of evaluation episodes & 100 \\
    \bottomrule
\end{tabular}
\caption{Evaluation details}
\label{tab:appendix_pi05_evaluation}
\end{subtable}

\caption{\textbf{Hyperparameter details for $\pi_{0.5}$ based methods.}}
\label{tab:appendix_pi05_details}
\end{table}

\subsection{Appendix E: Detailed Evaluation Results}
\label{sec:appendix_e}

Table~\ref{tab:main_results_combined} and Table~\ref{tab:fastwam_results_combined} reports the full set of quantitative results across all four manipulation tasks under both static (PosCons) and dynamic (PosCons + VelCons) constraint regimes, with all four evaluation metrics (Success Rate, Step Safety Rate, LDLJ, MMD). Note that all constrained methods deterministically achieve $100\%$ SSR at deployment; the SSR column is retained mainly to show the gap against the unconstrained baseline. Under the more challenging dynamic constraint regime, ActSafeGuard consistently attains the best or second-best performance across smoothness (LDLJ) and distributional alignment (MMD) while retaining the highest task success rate simultaneously.

\begin{table*}[htbp]
\centering
\caption{\textbf{Quantitative task success performance, safety, and trajectory quality across four manipulation tasks.} We report Success Rate (SR, \% $\uparrow$), Step Safety Rate (SSR, \% $\uparrow$), Log-Determinant of Lower-bound Jacobian (LDLJ $\uparrow$), and Maximum Mean Discrepancy (MMD $\downarrow$) under both static Position Constraints (PosCons) and combined Position + Velocity Constraints (PosCons + VelCons). Bold numbers indicate the best performance among constrained methods.}
\label{tab:main_results_combined}
\footnotesize 
\setlength{\tabcolsep}{3.5pt} 
\begin{tabular}{llcccc c cccc}
\toprule
\multirow{2}{*}{\textbf{Task}} & \multirow{2}{*}{\textbf{$\pi_{0.5}$ backbone}} & \multicolumn{4}{c}{\textbf{PosCons}} & & \multicolumn{4}{c}{\textbf{PosCons + VelCons}} \\
\cmidrule(r{4pt}){3-6} \cmidrule(l{4pt}){8-11}
& & \textbf{SR (\%)} $\uparrow$ & \textbf{SSR (\%)} $\uparrow$ & \textbf{LDLJ} $\uparrow$ & \textbf{MMD} $\downarrow$ & & \textbf{SR (\%)} $\uparrow$ & \textbf{SSR (\%)} $\uparrow$ & \textbf{LDLJ} $\uparrow$ & \textbf{MMD} $\downarrow$ \\
\midrule
\multirow{5}{*}{\textit{lift pot}} 
& Unconstrained Baseline & 100 & 83.59 & -14.578 & 0.01188 & & 100 & 82.85 & -14.578 & 0.01188 \\
& Projection             & \textbf{100} & 100 & -14.400 & 0.01195 & & \textbf{100} & 100 & \textbf{-13.998} & 0.01213 \\
& Truncation             & 99  & 100 & -14.487 & 0.01184 & & 99  & 100 & -14.472 & 0.01184 \\
& GaugeFlow              & 100 & 100 & -16.252 & \textbf{0.00564} & & N/A & N/A & N/A & N/A \\
& \textbf{ActSafeGuard (Ours)} & \textbf{100} & 100 & \textbf{-14.147} & 0.00627 & & \textbf{100} & 100 & -15.136 & \textbf{0.00995} \\
\midrule
\multirow{5}{*}{\textit{place shoe}} 
& Unconstrained Baseline & 90  & 20.43 & -18.205 & 0.00105 & & 90  & 16.45 & -18.205 & 0.00105 \\
& Projection             & 90  & 100   & -18.035 & 0.00149 & & 84  & 100   & -18.457 & 0.00245 \\
& Truncation             & \textbf{91}  & 100   & -18.161 & \textbf{0.00115} & & 90  & 100   & -18.201 & \textbf{0.00115} \\
& GaugeFlow              & 82  & 100   & -19.738 & 0.00510 & & N/A & N/A & N/A & N/A \\
& \textbf{ActSafeGuard (Ours)} & 90  & 100   & \textbf{-18.179} & 0.00348 & & \textbf{98}  & 100   & \textbf{-17.171} & 0.00242 \\
\midrule
\multirow{5}{*}{\textit{hanging mug}} 
& Unconstrained Baseline & 21  & 64.62 & -23.353 & 0.00219 & & 21  & 57.38 & -23.353 & 0.00219 \\
& Projection             & 21  & 100   & -22.994 & 0.00242 & & 26   & 100     & -22.351 & 0.00292 \\
& Truncation             & 27  & 100   & -22.813 & \textbf{0.00223} & & 27   & 100     & -22.796 & \textbf{0.00223} \\
& GaugeFlow              & 25   & 100     & -22.995 & 0.00265 & & N/A & N/A & N/A & N/A \\
& \textbf{ActSafeGuard (Ours)} & \textbf{34}  & 100   & \textbf{-21.509} & 0.00249 & & \textbf{31}  & 100   & \textbf{-21.996} & 0.00251 \\
\midrule
\multirow{5}{*}{\textit{place empty cup}} 
& Unconstrained Baseline & 90  & 37.04 & -17.235 & 0.01304 & & 90  & 35.69 & -17.235 & 0.01304 \\
& Projection             & 88  & 100   & -17.085 & 0.01352 & & 95  & 100   & -16.523 & 0.01350 \\
& Truncation             & 92  & 100   & -16.921 & 0.01305 & & 91  & 100   & -16.926 & 0.01256 \\
& GaugeFlow              & 93  & 100   & -17.165 & 0.01050 & & N/A & N/A & N/A & N/A \\
& \textbf{ActSafeGuard (Ours)} & \textbf{97}  & 100   & \textbf{-16.197} & \textbf{0.00960} & & \textbf{97}  & 100   & \textbf{-16.491} & \textbf{0.00913} \\
\bottomrule
\end{tabular}
\end{table*}

\begin{table*}[htbp]
\centering
\caption{\textbf{Quantitative task success performance, safety, and trajectory quality across four manipulation tasks.} We report Success Rate (SR, \% $\uparrow$), Step Safety Rate (SSR, \% $\uparrow$), Log-Determinant of Lower-bound Jacobian (LDLJ $\uparrow$), and Maximum Mean Discrepancy (MMD $\downarrow$) under both static Position Constraints (PosCons) and combined Position + Velocity Constraints (PosCons + VelCons). Bold numbers indicate the best performance among constrained methods.}
\label{tab:fastwam_results_combined}
\footnotesize
\setlength{\tabcolsep}{3.5pt}
\begin{tabular}{llcccc c cccc}
\toprule
\multirow{2}{*}{\textbf{Task}} & \multirow{2}{*}{\textbf{Fast-WAM backbone}} 
& \multicolumn{4}{c}{\textbf{PosCons}} & 
& \multicolumn{4}{c}{\textbf{PosCons + VelCons}} \\
\cmidrule(r{4pt}){3-6} \cmidrule(l{4pt}){8-11}
& & \textbf{SR (\%)} $\uparrow$ & \textbf{SSR (\%)} $\uparrow$ & \textbf{LDLJ} $\uparrow$ & \textbf{MMD} $\downarrow$
& & \textbf{SR (\%)} $\uparrow$ & \textbf{SSR (\%)} $\uparrow$ & \textbf{LDLJ} $\uparrow$ & \textbf{MMD} $\downarrow$ \\
\midrule
\multirow{5}{*}{\textit{lift pot}}
& Unconstrained Baseline & 100 & 15.12 & -21.051 & 0.64223 & & 100 & 13.68 & -21.051 & 0.64223 \\
& Projection             & \textbf{100} & 100 & \textbf{-15.577} & 0.64256 & & \textbf{100} & 100 & \textbf{-22.570} & 0.64256 \\
& Truncation             & \textbf{100} & 100 & -15.858 & 0.64319 & & 98 & 100 & -22.868 & \textbf{0.63280} \\
& GaugeFlow              & \textbf{100} & 100 & -15.939 & \textbf{0.61835} & & N/A & N/A & N/A & N/A \\
& \textbf{ActSafeGuard (Ours)} & 97 & 100 & -15.858 & 0.63021 & & \textbf{100} & 100 & -21.586 & 0.64256 \\
\midrule
\multirow{5}{*}{\textit{place shoe}}
& Unconstrained Baseline & 93 & 5.86 & -18.792 & 0.44594 & & 93 & 3.79 & -18.792 & 0.44594 \\
& Projection             & 85 & 100 & -18.983 & 0.45676 & & 85 & 100 & -19.139 & 0.45676 \\
& Truncation             & 94 & 100 & -18.387 & 0.45645 & & 80 & 100 & -18.277 & \textbf{0.45012} \\
& GaugeFlow              & 80 & 100 & -18.272 & 0.40675 & & N/A & N/A & N/A & N/A \\
& \textbf{ActSafeGuard (Ours)} & \textbf{95} & 100 & \textbf{-17.898} & \textbf{0.36872} & & \textbf{90} & 100 & \textbf{-18.175} & 0.46872 \\
\midrule
\multirow{5}{*}{\textit{hanging mug}}
& Unconstrained Baseline & 42 & 39.69 & -22.916 & 0.40783 & & 42 & 35.24 & -22.916 & 0.40783 \\
& Projection             & 35 & 100 & -22.894 & 0.41325 & & 40 & 100 & -23.201 & 0.40782 \\
& Truncation             & \textbf{45} & 100 & -22.751 & 0.41348 & & \textbf{50} & 100 & -23.051 & 0.53550 \\
& GaugeFlow              & 39 & 100 & -22.484 & 0.45024 & & N/A & N/A & N/A & N/A \\
& \textbf{ActSafeGuard (Ours)} & \textbf{45} & 100 & \textbf{-22.390} & \textbf{0.36681} & & 45 & 100 & \textbf{-22.711} & \textbf{0.39186} \\
\midrule
\multirow{5}{*}{\textit{place empty cup}}
& Unconstrained Baseline & 98 & 31.27 & -17.971 & 0.62598 & & 98 & 27.33 & -18.381 & 0.62598 \\
& Projection             & 82 & 100 & -23.799 & 0.96232 & & 80 & 100 & -19.245 & 0.69076 \\
& Truncation             & 88 & 100 & -23.018 & 0.96232 & & 90 & 100 & -19.852 & \textbf{0.61108} \\
& GaugeFlow              & 90 & 100 & -21.961 & 0.72542 & & N/A & N/A & N/A & N/A \\
& \textbf{ActSafeGuard (Ours)} & \textbf{95} & 100 & \textbf{-17.290} & \textbf{0.62312} & & \textbf{93} & 100 & \textbf{-18.108} & 0.62312 \\
\bottomrule
\end{tabular}
\end{table*}




\subsection{Appendix F: Real-Robot Deployment}

\begin{figure}
    \centering
    \includegraphics[width=0.8\linewidth]{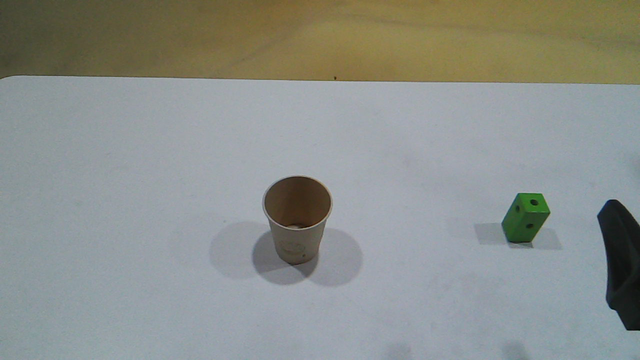}
    \caption{Setup snapshot of pick green cube task.}
    \label{fig:real_setup_pick}
\end{figure}

\paragraph{Pick green cube.}
The instruction for this task is: ``Pick up the green block and place it into the paper cup.'' In this task, only the right arm is used for manipulation, while the left arm remains stationary. We collect $50$ successful demonstrations using the ALOHA teleoperation interface and use them as the training dataset. The policy is trained to output absolute joint-position actions. We impose PosCons on the right-arm joint outputs, and the constraint bounds are obtained using the procedure described in Appendix~A. The lower bound is
[-0.155,\; 0.004,\; -1.034,\; -1.027,\; -1.204,\; -0.731],
and the upper bound is
[0.854,\; 1.747,\; -0.001,\; 0.803,\; 0.364,\; 1.012].
Figure~\ref{fig:real_setup_pick} shows the real-robot setup for this task.

\begin{figure}
    \centering
    \includegraphics[width=0.8\linewidth]{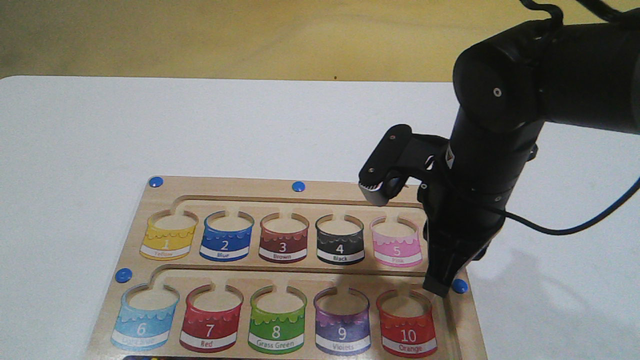}
    \caption{Setup snapshot of guide the ball task.}
    \label{fig:real_setup_guide}
\end{figure}

\begin{figure}
    \centering
    \includegraphics[width=\linewidth]{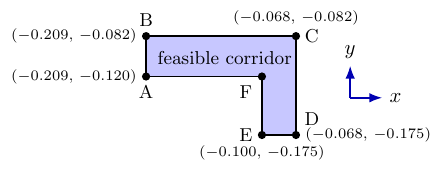}
    \caption{Constraints defined in guide the ball task. }
    \label{fig:real_cons_guide}
\end{figure}

\paragraph{Guide the ball.}
The instruction for this task is: ``Guide the green ball along the track into the green bucket.'' Similar to the pick-green-cube task, only the right arm is used for manipulation and the left arm remains stationary. We collect $50$ successful demonstrations using the ALOHA teleoperation interface and use them as the training dataset. In this task, the policy outputs a 7-DoF end-effector pose for every arm, consisting of a 3-DoF position and a 4-DoF quaternion orientation. The constraint is defined as an L-shaped feasible corridor in the end-effector position space. Figure~\ref{fig:real_setup_guide} shows the real-robot setup and Figure~\ref{fig:real_cons_guide} shows the corresponding feasible region.
In the guide-the-ball task, the constraints are imposed on the $x$ and $y$ dimensions of the end-effector position. Notably, the feasible region in this task is non-convex. ActSafeGuard can naturally handle this non-convex constraint because it only requires a ray-shooting computation along the predicted action direction, and the non-convex geometry does not introduce additional optimization complexity. In contrast, other constraint-handling approaches are much harder to apply in this setting. QP-based projection or CBF methods do not directly extend to non-convex feasible sets, and gauge-map-based methods are also not applicable to such non-convex domains.

\subsection{Appendix G: Inference Algorithm}
Algorithm~\ref{alg:inference} shows the inference process of ActSafeGuard.

\begin{algorithm}[h]
\caption{ActSafeGuard Inference}\label{alg:inference}
\begin{algorithmic}[1]
\REQUIRE Current observation $\mathbf{o}_t$, language instruction $l$
\STATE Encode context: $\mathbf{e}_t = f_\theta(\mathbf{o}_t, l)$
\STATE Sample initial noise $\mathbf{x}_0 \sim p(\mathbf{x}_0)$ s.t. $\mathbf{x}_0 \in \mathcal{C}(\mathbf{o}_t)$
\FOR{$k = 0$ \TO $N-1$}
    \STATE Set flow time $\tau_k = k/N$
    \STATE Predict baseline velocity $\mathbf{v}_\phi \leftarrow \mathbf{v}_\phi(\mathbf{x}_k, \tau_k, \mathbf{e}_t)$
    \STATE Compute direction vector $\mathbf{d}_\phi = \mathbf{v}_\phi / N$
    \STATE Compute scale $s_\phi$ using Eq (7) and (8)
    \STATE Apply safe update $\Delta_\phi = s_\phi \mathbf{d}_\phi$
    \STATE Update feasible system state: $\mathbf{x}_{k+1} = \mathbf{x}_k + \Delta_\phi$
\ENDFOR
\RETURN $\mathbf{x}_N$ (Executable safe action chunk)
\end{algorithmic}
\end{algorithm}


\end{document}